\documentclass[STIX1COL]{WileyNJD-v2}
\articletype{}%

\received{Date Month Year}
\revised{Date Month Year}
\accepted{Date Month Year}

\usepackage{xspace}
\usepackage{upgreek}
\usepackage{cite}
\usepackage{amsmath,amsfonts}
\usepackage{varwidth}
\usepackage[caption=false,font=footnotesize]{subfig}
\usepackage{ragged2e}
\usepackage{booktabs}
\usepackage[linesnumbered, ruled, vlined]{algorithm2e}
\usepackage{url}
\usepackage{soul}
\DeclareUnicodeCharacter{2010}{-}

\usepackage{pifont}
\newcommand{\cmark}{\ding{51}}%
\newcommand{\xmark}{\ding{55}}%

\definecolor{ForestGreen}{RGB}{34,139,34}
\usepackage{tikz}
\newcommand*\circled[1]{\tikz[baseline=(char.base)]{
		\node[shape=circle,draw,inner sep=1pt] (char) {#1};}}
\newcommand{\tikzcircle}[2][green,fill=ForestGreen]{\tikz[baseline=-0.5ex]\draw[#1,radius=#2] (0,0) circle ;}

\newcommand{\theterm}{Perception~Simplex\xspace}%
\newcommand{\ps}{$\mathcal{PS}$\xspace}%
\newcommand{\lidar}{LiDAR\xspace}

\newcommand{\eg}{{\it e.g.,}\xspace}

\newcommand{\ie}{{\it i.e.,}\xspace}

\newcommand{\ci}{{\it (i) }}
\newcommand{\cii}{{\it (ii) }}
\newcommand{\ciii}{{\it (iii) }}
\newcommand{\civ}{{\it (iv) }}
\newcommand{\cv}{{\it (v) }}

\newcommand{\ca}{{\it (a) }}
\newcommand{\cb}{{\it (b) }}

\newcommand{\vs}{\textit{vs}}

\begin{document}

\title{ 
Perception Simplex: Verifiable Collision Avoidance in Autonomous Vehicles Amidst Obstacle Detection Faults}

\author[1]{Ayoosh Bansal}

\author[2]{Hunmin Kim}

\author[1]{Simon Yu}

\author[1]{Bo Li}

\author[1]{Naira Hovakimyan}

\author[3]{Marco Caccamo}

\author[1]{Lui Sha}

\authormark{A. Bansal, H. Kim, S. Yu, B. Li, N. Hovakimyan, M. Caccamo and L. Sha}
\titlemark{Perception Simplex: Verifiable Collision Avoidance in Autonomous Vehicles Amidst Obstacle Detection Faults.}

\address[1]{\orgname{University of Illinois Urbana-Champaign}, \orgaddress{\state{Illinois}, \country{USA}}}

\address[2]{\orgname{Mercer University}, \orgaddress{\state{Georgia}, \country{USA}}}

\address[3]{\orgname{Technical University of Munich}, \orgaddress{\state{Munich}, \country{Germany}}}

\corres{Ayoosh Bansal \\ \email{ayooshb2@illinois.edu}}

\fundingInfo{
    \justifying
    \noindent
    The material presented in this paper is based upon work supported by
    the National Science Foundation (NSF) under grant no. CNS 1932529, ECCS 2020289,
    the Air Force Office of Scientific Research (AFOSR) under grant no. \#FA9550-21-1-0411,
    the National Aeronautics and Space Administration (NASA) under grant no. 80NSSC22M0070, 80NSSC20M0229, AWD-000577-G1,
    and University of Illinois Urbana-Champaign under grant no. STII-21-06.
    Marco Caccamo was supported by an Alexander von Humboldt Professorship
    endowed by the German Federal Ministry of Education and Research.
    Any opinions, findings, conclusions, or recommendations expressed in
    this publication are those of the authors and don't necessarily
    reflect the views of the sponsors.%
}

\abstract[Abstract]{

Advances in deep learning have revolutionized cyber-physical applications, including the development of Autonomous Vehicles.
However, real-world collisions involving autonomous control of vehicles have raised significant safety concerns regarding the use of Deep Neural Networks (DNN) in safety-critical tasks, particularly Perception.
The inherent unverifiability of DNNs poses a key challenge in ensuring their safe and reliable operation.

In this work, we propose \theterm (\ps), a fault-tolerant application architecture designed for obstacle detection and collision avoidance.
We analyze an existing \lidar-based classical obstacle detection algorithm to establish strict bounds on its capabilities and limitations.
Such analysis and verification have not been possible for deep learning-based perception systems yet.
By employing verifiable obstacle detection algorithms, \ps identifies obstacle existence detection faults in the output of unverifiable DNN-based object detectors.
When faults with potential collision risks are detected, appropriate corrective actions are initiated.
Through extensive analysis and software-in-the-loop simulations, we demonstrate that \ps provides predictable and deterministic fault tolerance against obstacle existence detection faults, establishing a robust safety guarantee.

}

\keywords{autonomous vehicles, software reliability, fault tolerance, cyber-physical systems, obstacle detection}

\maketitle

\renewcommand\thefootnote{}

\footnotetext{\textbf{Abbreviations:} \ps, perception simplex; AV, autonomous vehicles; DNN, deep neural networks; CPS, cyber-physical systems.}
\footnotetext{\textbf{Extension from Prior Work:}
This work is a substantially revised and extended version of a paper presented at ISSRE 2022~\cite{bansal2022verifiable}.
The main difference is that while the prior work presented the \textit{detectability} model for the verifiable obstacle detection algorithm, this work develops the system architecture to utilize the verifiable capabilities and provide deterministic fault tolerance.
The major differences are
\ci   the design, analysis and implementation for \ps~($\S$\ref{sec:ps});
\cii  evaluation of \ps using software-in-the-loop simulation based on industrial simulators~($\S$\ref{sec:eval_ps});
\ciii alleviate a constraint of the \textit{detectability} model by accommodating for
        \lidar beam energy attenuation in presence of air impediments~($\S$\ref{sec:lidar_attenuate});
\civ  expanded related work to address the recent state of the art and the new contributions~($\S$\ref{sec:relatedwork});
\cv other editorial changes to most sections, especially the abstract, introduction, scope and motivation, and conclusion.}

\renewcommand\thefootnote{\arabic{footnote}}
\setcounter{footnote}{0}

\section{Introduction}
\label{sec:introduction}

Autonomous Vehicles (AV) hold great promise in enhancing road safety and potentially saving lives. However, despite the rapid progress and widespread deployment of AV prototypes, the road to achieving collision-free AV remains uncertain~\cite{dirty_dozen}. Instances of fatal collisions involving autonomous control of vehicles have raised significant concerns~\cite{
	crash_tesla_2016_1_20_china,crash_ntsb_tesla_2016_5_7,
	crash_ntsb_uber_2018_3_18,crash_ntsb_tesla_2018_3_23,crash_tesla_2018_4_29_japan,
	crash_ntsb_tesla_2019_3_1,crash_tesla_2019_4_25,crash_tesla_2019_8_24,crash_tesla_2019_12_29_1,crash_tesla_2019_12_29_2,
	crash_tesla_2020_5_29,
	crash_tesla_2021_5_5,crash_tesla_2021_7_26,crash_nio_2021_8_12,
	crash_tesla_2022_5_16,crash_tesla_2022_5_17,crash_tesla_2022_7_6,crash_tesla_2022_7_24,crash_tesla_2022_8_26,
	crash_tesla_2023_2_18}.
These incidents underscore the continued need for research and development efforts, aimed at enhancing the safety and reliability of AV.

AV depend on deep learning techniques to perform various tasks, including perceiving the vehicle's environment. However, Deep Neural Networks (DNN) operate by extrapolating correlations learned from training data, which inherently renders them unverifiable. As a result, their behavior becomes unpredictable and unreliable, rendering them unsuitable for deployment in safety-critical tasks~\cite{heaven2019deep,huang2020survey,pereira2020challenges,willers2020safety,tambon2022certify}. Safety-critical software necessitates logical analysis and verification, requirements that current DNN solutions are not yet equipped to fulfill. The challenges are further compounded by the design of autonomous driving systems, where mission-critical tasks (navigation) and safety-critical tasks (collision avoidance) often overlap without clear delineation. This lack of separation leads to ill-formed optimization objectives for the system, and the safety problem becomes entangled and convoluted.

\begin{figure}[t]
	\centering
	\includegraphics[width=.5\linewidth]{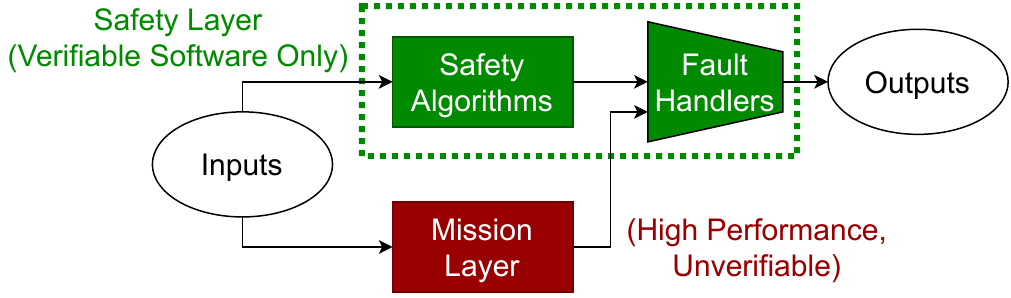}
	\caption{\label{fig:ps_overview}
		\theterm design.
	}
\end{figure}

Hence, while DNN-based solutions are essential for fulfilling the mission of an AV, they also pose significant safety challenges.
This research proposes a solution to this apparent dilemma by introducing \textbf{\theterm} (\ps), which enables deterministic fault tolerance against perception faults in AV.
To achieve this, we adopt a decoupled approach, similar to previous solutions in control systems~\cite{simplex_original,simplex2}.
The \ps architecture consists of a safety layer and a mission layer.
The safety layer incorporates logically analyzable and verifiable algorithms that focus on detecting faults and ensuring minimal functionality for safety-critical requirements.
On the other hand, the mission layer houses the sophisticated self-driving software, \eg Apollo~\cite{apollo} or Autoware~\cite{autoware}, which is responsible for fulfilling all aspects of the AV's mission while trying to maintain safety.
The proposed procedure within this architecture involves:
    \ci identifying the safety-critical requirements,
    \cii utilizing verifiable algorithms to detect faults in meeting these requirements, and
    \ciii continuously monitoring for faults at runtime and initiating corrective actions when necessary.
An overview of the \ps architecture is presented in Figure~\ref{fig:ps_overview}.

This work presents a brief survey of real-world fatal collisions associated with autonomous control of vehicles.
The survey reveals that the most prevalent type of fault involved in these collisions is related to the detection of obstacle existence.
To address this critical issue, we apply the \ps framework, which offers deterministic fault tolerance specifically targeting these obstacle existence detection faults.
Additionally, we design \ps with a flexible and modular structure, enabling future expansions to address other perception faults.
To address obstacle existence faults with \ps, we first carefully reduce the requirements for obstacle existence detection for collision avoidance.
By focusing on only the essential elements necessary for this specific task, we eliminate several features of mission-critical object detection that are deemed unnecessary.
This streamlined approach allows us to fulfill the task of obstacle existence detection while minimizing complexity.
Furthermore, we conduct a detailed analysis of an existing \lidar-based geometric algorithm called Depth Clustering~\cite{bogoslavskyi16iros, bogoslavskyi17pfg}.
Through this analysis, we establish a comprehensive model, named the \textit{detectability} model, which outlines the algorithm's capabilities and limitations.
The Depth Clustering algorithm is then integrated into the \ps pipeline, where the \textit{detectability} model serves as the foundation for ensuring safety guarantees in collision avoidance scenarios.
The key \textbf{contributions} of this work are:%
\begin{itemize}

 \item Reduce the complexity of safety-critical obstacle existence detection by carefully eliminating unnecessary object detection features, including object type classification ($\S$\ref{sec:minimal_ca}).

 \item Establish a \textit{detectability} model for an existing \lidar-based geometric obstacle detection algorithm, providing human perceptible bounds on capabilities and limitations ($\S$\ref{sec:model}).

 \item Design, analyses, and implementation of the \theterm (\ps) framework to achieve deterministic fault tolerance specifically targeting the perception fault of obstacle existence detection ($\S$\ref{sec:ps}).

 \item Evaluation of the verifiable algorithm using real sensor data~\cite{sun2020scalability}~($\S$\ref{sec:eval_vod})
    and software-in-the-loop simulation \cite{apollo,lgsvl,apollo_lgsvl} demonstrating \ps's response in  obstacle existence detection fault scenario~($\S$\ref{sec:eval_ps}).\footnote{\url{https://github.com/CPS-IL/perception_simplex}}
\end{itemize}

\section{Scope and Motivation}
\label{sec:motivation}
\label{sec:scope}

\begin{table}[t]
    \centering
    \caption{\label{tab:crash_survey}%
        Survey of fatal traffic collisions involving autonomous control of vehicles.}
    \begin{tabular}{c|l}
        \toprule
        Reference
        &   Autonomy Response\tnote{$^\dagger$}
        \\ \midrule

        \cite{crash_ntsb_tesla_2016_5_7}
        & A truck in the path of the vehicle was not detected and no evasive actions were initiated.
        \\ \midrule

        \cite{crash_ntsb_uber_2018_3_18} 
        & Changing the classification of a pedestrian led to the perception system ignoring their existence.
        \\ \midrule

        \cite{crash_ntsb_tesla_2018_3_23} & A crash attenuator in the path of the vehicle was not detected.
        \\ \midrule

        \cite{crash_ntsb_tesla_2019_3_1} & A white semi-trailer in the path of the vehicle was not detected.
        \\ \bottomrule
    \end{tabular}
    \begin{tablenotes}
        \footnotesize \centering
        \item[$^\dagger$] The comments brief the driver assist system's response during the incident and are not necessarily the causal faults for the incident.
    \end{tablenotes}    
\end{table}

Obstacle existence detection fault, \ie False Negative (FN) in perception is a grave safety concern~\cite{yang2021introspective}.
A survey of fatal collisions involving AV (Table~\ref{tab:crash_survey}) points to recurring FN errors.
Other fatalities involving AV~\cite{
	crash_tesla_2016_1_20_china,
	crash_tesla_2018_4_29_japan,
	crash_tesla_2019_4_25,crash_tesla_2019_8_24,crash_tesla_2019_12_29_1,crash_tesla_2019_12_29_2,
	crash_tesla_2020_5_29,
	crash_tesla_2021_5_5,crash_tesla_2021_7_26,crash_nio_2021_8_12,
	crash_tesla_2022_5_16,crash_tesla_2022_5_17,crash_tesla_2022_7_6,crash_tesla_2022_7_24,crash_tesla_2022_8_26,
	crash_tesla_2023_2_18},
excluded from Table~\ref{tab:crash_survey} due to the unavailability of public investigation reports, seem to follow this pattern.
These underlying safety concerns are the primary challenge in adopting complete autonomy~\cite{nhtsa_investigation,penmetsa2021effects,li2022safety,jing2023listen}.
Therefore this work focuses on providing deterministic fault tolerance against obstacle existence detection faults.
Causes, mechanisms, and adversarial inducement of such faults is an active area of research~\cite{tu2020physically,miller2022s}.
However, these causes are out of the scope of this work, therefore, to evaluate \ps the faults are injected via instrumentation, as described in Section~\ref{sec:eval_setup}.

Analyzability and verifiability are the crucial components of the certification process of safety-critical \mbox{systems~\cite{feiler2013four,heimdahl2016software}}.
Verifiable algorithms, where the \textit{causality} between the input parameters and the algorithm result can be established, are inherently suitable for safety-critical applications.
This is in contrast to object detection DNN, trained using supervised learning, which effectively captures \textit{correlations} between training input and labels.
Consider the following simple example where $y$ is obstacle height, $x$ is obstacle distance from the AV, $a$ and $b$ are constant parameters based on \lidar properties:
\begin{equation}
    y \geq ax + b \label{eq:ideal}
\end{equation}
If we want to establish that \eqref{eq:ideal} is the \textit{detectability} model, \ie when the condition in \eqref{eq:ideal} is met, obstacles are always detected by the AV,
the following must be defined:

    {\it \textbf{Requirements}}: A definition of minimal requirements for what it means to \textit{detect} an obstacle ($\S$\ref{sec:minimal_ca}).

    {\it \textbf{Constraints}}: A set of well-defined constraints that must be met for the model to be applicable ($\S$\ref{sec:paramconstraints}).

    {\it \textbf{Verification}}: A deterministic analysis verifying the \textit{detectability} model ($\S$\ref{sec:model}).

Let's assume that an AV safety standard requires that the AV be able to detect all obstacles of a minimum height of $10$~$cm$.
Further, let's assume all vehicles and structures on the road are also mandated to be taller than this minimum height by a road safety rule.
Using \lidar and AV parameters to determine $a$ and $b$, the \textit{detectability} model \eqref{eq:ideal} can be used to determine the minimum distance at which such obstacles can be detected.
This minimum distance, in conjunction with the braking capability of the AV, can then be used to determine the max speed at which the AV can safely travel.
This example, admittedly simple, shows how a verified \textit{detectability} model can bring together road safety rules, AV parameters, and AV safety policies to provide deterministic collision safety.
We show in this work that such velocity limitations can be derived, as in Theorem~\ref{lem:safe},
  guaranteeing collision avoidance when complex mission-critical object detectors suffer obstacle existence detection faults.

It is assumed that all communication in the \ps system is real-time and lossless.
Data shared between layers is assumed to be always fresh and valid for its use.
Temporal safety \ie real-time execution and communication are important to \ps design, though have been the focus of prior works~\cite{menard2020achieving,baron2021lett,mirosanlou2021duetto}.
This work presents the functional design and safety guarantees of \ps.

While we limit this work to obstacle existence detection faults due to their importance and prevalence in real-world fatal collisions,
  \ps is designed to be extensible,
  to address other perception faults by utilizing verifiable capabilities to fulfill other safety-critical perception tasks,
  as exemplified by the handling of obstacle existence detection faults in this work.
Furthermore, while the \ps is applied to autonomous ground vehicles in this work,
  the procedure for verifiable mitigation of faults in complex DNN,
  can be applied to other safety-critical systems reliant on DNN solutions.
\section{Related Work}
\label{sec:relatedwork}

\textbf{Redundancy}
is a crucial concept for designing reliable safety-critical cyber-physical systems (CPS).
Full system replication enables a system to tolerate and correct random or systemic faults that an individual system cannot~\cite{lala1994architectural, 963314, isermann2002fault, baleani2003fault, kim2005design, rooks2005duo}.
However, computational redundancy does not address faults shared by all replicas \textit{i.e.} algorithmic or implementation faults~\cite{6629559, 7823109}.
Functional or analytical redundancy for fault-tolerant control~\cite{blanke2006diagnosis},
  addresses this by using different systems with different implementations, inputs, capabilities, or algorithms to fulfill the same function \eg GPS and inertial navigation.

\textbf{Simplex}
architecture~\cite{simplex_original,simplex0,simplex2} and
  its evolutionary variations~\cite{kwon2018approach,phan2020neural,wang2018rsimplex}, including \ps,
  are a form of asymmetric functional redundancy.
This architecture safeguards systems that are controlled by a \textit{high-performance} but potentially unreliable controller,   switching to a reliable, \textit{high-assurance} controller when needed, in order to ensure the safety of the system.
  Although solutions have been proposed to tackle control~\cite{mao2023sl1} and planning~\cite{musau2022using} in autonomous vehicles
  and to account for uncertainty in position and velocity measurements~\cite{bernhard2022risk},
  these approaches do not directly address the critical issue of gross lapses in perception, \eg obstacle existence detection faults.  
\ps is the first to propose a deterministic simplex implementation for fault tolerance against obstacle existence detection faults,
  by leveraging verifiable solutions for obstacle detection.
\ps is a modular solution designed to be open to future integration of other solutions for measurement uncertainty, planning, and control.

\textbf{DNN verification},
even for simple properties, is an NP-complete problem~\cite{katz2017reluplex}.
Gharib \etal~\cite{gharib2018safety} describe the need and current lack of verification methods for machine learning components used in safety-critical applications.
Albarghouthi~\cite{albarghouthi2021introduction} described the challenges for neural network verification, including scale, complexity, and dynamism of the environment, all applicable for their use in Autonomous Driving.
Liu~\etal~\cite{liu2021algorithms} survey existing verification methods and classify the verification methods into three categories: reachability, optimization, and search,
  identifying tradeoffs between scale and completeness of the verification methods.
Even when the deep networks can be verified, it is done so for small input sets only~\cite{tran2020verification}.
Improving the input coverage by exploring around available inputs by adding adversarial or context-specific distortions~\cite{huang2017safety, tian2018deeptest} is insufficient in the face of the vastness of the input sets for real-world problems, like perception.

\textbf{Classical Obstacle Detection}
approaches have been developed using various sensors.
Long-range obstacle detection based on stereo cameras was proposed by Pinggera~\etal~\cite{pinggera2015high}.
\lidar sensors have shown incredible promise for their use in autonomous driving~\cite{li2020lidar}.
This has, in turn, accelerated the improvements in \lidar sensor technology, increasing the range, scanning frequency, and resolution of the sensor~\cite{roriz2021automotive}.
\lidar sensors suffer a linear error growth with distance as compared to quadratic for stereo camera~\cite{wang2019pseudo}.
These factors made the \lidar sensor the natural choice for this work.
Various \lidar-based geometric algorithms were considered as part of this work~\cite{himmelsbach2010fast, korchev2013real, asvadi2015detection, zermas2017fast}.
Each algorithm has a similar flow; identifying points on the ground vs. obstacles, followed by clustering, segmentation, and/or classification.
Depth Clustering~\cite{bogoslavskyi16iros, bogoslavskyi17pfg} is chosen as the primary example due to its deterministic explainable behavior, flexible parameters to optimize tradeoffs, and public availability of an efficient implementation with extremely low runtime of $40$ $ms$ on an embedded Jetson Xavier platform, using a single CPU core only, for a point cloud with $169,600$ points~\cite{chen2021lidar}.

\textbf{Sensor Fusion}
is used in AV~\cite{realpe2016multi} to alleviate limitations of individual sensor types.
However, the intermingling of safety and performance optimizations, limitations of the algorithms used to interpret sensor inputs,
and unforeseen interactions between features unnecessary for safety and optimizations for performance have resulted in perception lapses and collisions~\cite{crash_ntsb_tesla_2016_5_7,crash_ntsb_uber_2018_3_18,crash_ntsb_tesla_2018_3_23,crash_ntsb_tesla_2019_3_1}.
To address these issues, the \ps safety layer uses algorithms that have deterministic capabilities, constraints, and limits.
However verifiable algorithms using different sensor data types would help alleviate the constraints of individual algorithms.
The use of sensor fusion in \ps is left for future work.

\textbf{Safe Autonomous Driving}
has been an active area of research with varying approaches.
System architectures have been proposed that make parts of the AV software have better security and execution time predictability~\cite{burgio2017software,drive_os,mirosanlou2021duetto,chen2022parallel}.
While this addresses the response latency variance for the AV, it does not address the faults born out of complex algorithms in the software pipelines.
Cooperative approaches~\cite{chen2023safe} do not address the mixed autonomous and manually driven vehicle conditions.
Techniques to improve driving policies~\cite{yuan2020race,cao2022trustworthy}, path planning~\cite{lee2019collision,wang2019crash,tahir2019heuristic}, risk assessment~\cite{noh2017decision,noh2018decision,shin2018human,yu2019occlusion,li2021risk}, steering or braking control~\cite{seiler1998development,funke2016collision,he2019emergency,cheng2019longitudinal,hajiloo2020integrated} are still vulnerable to perception failures.
\ps addresses the perception failures first and is flexible to the inclusion of advancements in the mission layer.

\textbf{Summary.}
\ps integrates the design philosophies of various prior safety architectures for AV while providing targeted safety guarantees and the flexibility of incremental development making it practical and easy to include in existing AV software stacks.
Existing approaches for safe control can be housed within this architecture to expand the capabilities and fault handling of the safety layer.
This work provides a path to verifiable fulfillment of safety properties in AV, starting from Perception.

\section{Minimal Requirements}
\label{sec:minimal_ca}

In this section, we define the minimally sufficient, though not necessary and sufficient, requirements for safety-critical obstacle detection and collision avoidance.
These requirements define a minimized set of features of an obstacle that the AV perception system should perceive to establish the existence of the obstacle.
While some of the following observations are already well established, this work is the first to form a minimally sufficient set for safety-critical collision avoidance.
Further, some requirements like object classification are typically considered crucial for all aspects of perception in AV, however, we show that while it is valuable for mission-critical requirements, it is not for safety-critical collision avoidance.

\subsection{Classification}
\label{sec:classification_not_required}

We will use the mathematical problem formulation of reach-avoid control to justify why we do not need to classify obstacles for collision avoidance purposes.
The current section is motivated by~\cite{summers2010verification,summers2011stochastic} but we simplify the notations and the problem in a deterministic setting while the references consider stochastic problem formulations.
Let  $x_k \in X$ be the 2D-position of the vehicle at discrete time instance $k \geq 0$, where $X \subseteq {\mathbb{R}}^2$ is the set of the state space. The set $D \subset X$ is the destination, and the sets $O_i(k) \subset X$ for $i \in M_O \triangleq \{1,2,\cdots,M\}$ are obstacles, where $O_i(k)$ represents the $i^{th}$ obstacle and $M_O$ is the index set for the obstacles. The set $O_i(k)$ could be time-varying but it is assumed that the obstacles do not overlap with the destination, \ie $\cup_{i \in M_O} (D \cap O_i(k)) = \emptyset$ $\forall$ $k$. For brevity, further mentions of $O_i(k)$ are reduced to $O_i$.
Now, the reach-avoid control problem is to drive the vehicle to the destination $D$ while avoiding obstacles $O_i$ for $i \in M_O$ within finite time horizon $N$, given initial condition $x_0$.
The success of this mission can be characterized by the following index $\kappa$:
\begin{align}
	\kappa& \triangleq \textstyle \sum_{j=0}^N \big(\Pi_{i=1}^M \Pi_{t=0}^{j-1} \mathbf{1}_{X \setminus O_i}(x_t)\big) \mathbf{1}_D(x_j)\nonumber\\
	&=
	\left\{
	\begin{array}{cc}
		1, &  {\rm if \ } \exists j \in \{0,1,\cdots N\}: x_j \in D \wedge \forall t \in \{0,1, \cdots, j-1\}: x_t \in \cap_{i=1}^M X \setminus O_i\\
		0, & {\rm otherwise,}\\
	\end{array}
	\right.
	\label{eq:formulation0}
\end{align}
where $\mathbf{1}_S(\cdot): X \rightarrow 
\{0,1\}$ is the indicator function for a set $S$, and $\wedge$ is the logical AND. In short, $\kappa=1$ if and only if the objective is achieved. The index $\kappa$ in~\eqref{eq:formulation0} can be used as a cost function of the optimal control problem as in~\cite{summers2010verification,summers2011stochastic}. Therefore, one is required to evaluate the indicator functions in~\eqref{eq:formulation0}, which means that it is required to know the set $D$ and $X \setminus O_i$ for $i \in M_O$.

However, index $\kappa$ in~\eqref{eq:formulation0} can be equivalently formulated as
\begin{align}
	\kappa& = \textstyle \sum_{j=0}^N \big( \Pi_{t=0}^{j-1} \mathbf{1}_{X \setminus (\cup_{i \in M_O} O_i)}(x_t)\big) \mathbf{1}_D(x_j)
	\label{eq:formulation1}
\end{align}
and this expression can be used instead for the optimal reach-avoid control problem.
The formulation~\eqref{eq:formulation1} indicates that one could also address the control problem only with knowing $\cup_{i \in M_O} O_i$, but not individual obstacle sets $O_i$, \ie
one does not need to classify/distinguish individual obstacles for the reach-avoid control purpose.

It should be noted that classification adds valuable information that supports advanced features like predictive tracking, and maneuver planning and improves AV performance.
The argument here is only that object classification is not a necessity for obstacle avoidance and
therefore not a part of the minimal requirements for safety-critical obstacle detection.
\subsection{Collision Risk}
\label{sub:rrr}

Collision avoidance involves many factors, from perception to vehicular control.
While the dynamics and ethics of collision avoidance are complex~\cite{goodall2014ethical}, the safety-critical obstacle detection system is required to detect all obstacles that can potentially collide with the AV.
We utilize a physical model for collision risk from our prior work~\cite{9460196}, where
the potential risk of collision is determined by the overlap of the existence regions~\cite{schmidt2006research} of obstacles and AV within the AV's time to stop.

\subsection{Height}

\begin{figure*}[!t]
    \centering
    \includegraphics[width=\linewidth,keepaspectratio]{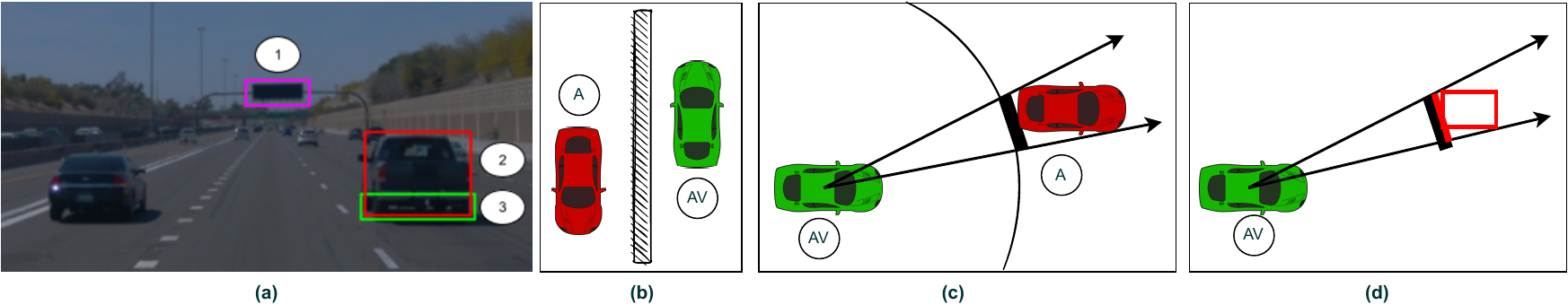}
    \caption{\label{fig:req_examples}
        Example scenes.
        (a) Role of obstacle height.
        (b) Distance and Occlusion of Obstacles.
        (c) Projection of Obstacles.
    }
\end{figure*}

The height of an obstacle is only useful in making a binary determination for collision avoidance, \ie
whether or not the obstacle is completely clear above the height of the AV.
For example, in Figure~\ref{fig:req_examples}~(a)  the overhead road sign \circled{1}  is completely above the AV, its exact height has no implication for collision avoidance.
The box \circled{2} contains valuable information that is required to identify the obstacle within the box to be a vehicle, however as discussed in Section~\ref{sec:classification_not_required}, such a recognition of the obstacle class is not a requirement for collision avoidance.
Thus for safety-critical collision avoidance \circled{3} contains as much relevant information as \circled{2}.
Therefore as long as an obstacle's height is not erroneously detected to be above and clear of the AV, we can simply consider the top or bird's eye view of the AV's surroundings.
While such a view of obstacles is not traditionally used in perception systems, however, path planning in AV, an inherently 2D problem, regularly uses this representation~\cite{ming2021survey}.

\subsection{Distance}
\label{sec:req_dist}

The distance to an obstacle must be accurately detected for the collision-free operation of the AV.
For collision avoidance, this distance is the minimum distance between the perimeters of the obstacle and the AV.
Many safety parameters like safe following distance, time to collision, and time to stop, are a function of the distance between the AV and obstacles~\cite{seiler1998development, cheng2019longitudinal},
and therein lies the difference in underestimation and overestimation error in distance detection.

\subsubsection{Underestimation Error}
When obstacle distances are underestimated, \ie obstacles are detected to be closer than they are, the performance of AV suffers,
however, there is no negative impact on direct collision avoidance.
An implicit effect of acceptance of all detection with a lower distance than the ground truth obstacle is that occluded obstacles become unnecessary to detect as long as the occluding obstacle is detected.
An example of this is Figure~\ref{fig:req_examples} (a) and (b).
A small wall exists to the left side of the AV in both images, separating traffic moving in the opposite direction to the AV.
If the wall is detected as an obstacle, the AV would not need to detect the vehicles across the wall to avoid colliding with them. Avoiding collisions with the occluding obstacle (wall here) means avoiding collisions with occluded obstacles.
Note that this only applies to completely occluded obstacles. Partial occlusion is discussed in Section~\ref{sec:req_coverage}.

\subsubsection{Overestimation Error}
\label{sub:dist_error_positive}
Overestimation of distance is a grave safety concern.
Obstacles detected to be further than they are, invalidate any safety decisions that would be based on this information.
Since it is unrealistic that an obstacle detection system would always have no distance overestimation error,
a strict bound on overestimation distance detection error is required.
With an established max error bound, any distance-based safety guarantee can assume this error is always present, maintaining the safety guarantee in the worst case.

\subsection{Projection}
\label{sec:req_coverage}

We determine a projection signifying a line on the road that the AV cannot cross without colliding with the obstacle for each obstacle.
Figure~\ref{fig:req_examples} (c) shows an example of this.
A circle is drawn with its center at the AV's sensor hub and radius equal to the closest point on the obstacle from AV.
The projection of the obstacle is then found as a line segment, tangent to the point where the above circle touches the obstacle.
This line segment is shown as a thick line segment in Figure~\ref{fig:req_examples} (c).
This line segment is a representation of the obstacle. Note that as new sensor inputs come in over time, the AV and obstacle move relative to each other, and the projection moves accordingly.
So the line segment only represents the obstacle in the current frame to make safety-critical decisions until the next sensor frame is received and processed.
With this minimal representation of the ground truth obstacles, containing only the minimal information about the obstacles as required for safety-critical collision avoidance, we can now determine when detected obstacles are sufficiently detected to avoid collisions.
\subsection{Coverage}

Each detected obstacle that meets the distance criteria ($\S$\ref{sec:req_dist}) and falls within the direction of an obstacle can now be used to determine if a ground truth obstacle is sufficiently covered by detected obstacles to avoid collisions with the ground truth obstacle.
Detected obstacles are projected on the projection of the ground truth obstacle to determine what parts of the ground truth projection are covered by the detected obstacles, as shown in Figure~\ref{fig:req_examples}~(d).
The projected detection must provide enough information to avoid collisions for an obstacle to be considered as detected.
As with distance ($\S$\ref{sec:req_dist}) this coverage may not be perfect but should have bounded errors.
A limited proportion of the ground truth projection must be covered. This is equivalent to the traditionally used intersection over union (IOU) limits.
It should be noted that the error margin of coverage is less important than that of the distance.
The minimum distance always tracks the distance between the closest points between the AV and the obstacle, changing with each sensor input frame.

\subsection{Summary}

In this section, we have discussed various features of obstacles a perception system may detect and detailed their relevance for safety-critical collision avoidance.
Many properties of obstacles that are considered critical parts of perception in AV
(\eg class, 3D dimensions, road sign information),
while indeed required for achieving the mission of autonomous driving, are not necessary for safety-critical collision avoidance.
In brief, an obstacle is considered detected for safety-critical obstacle avoidance if the following are accurately determined
\ca the distance of the obstacle from AV, within bounded error margins.
\cb a line on the road that the obstacle makes unsafe for the AV to cross.
An obstacle detection system that reliably meets the above requirements can be used to detect safety-critical faults in perception by more complex perception systems, thus providing fault detection and collision avoidance system.
The fusion of verifiable algorithms with DNN and the control decisions based on the verifiable algorithms have additional challenges that will be addressed in future works.

\section{Parameters and Constraints}
\label{sec:paramconstraints}

\subsection{\lidar  Parameters}
\label{sub:lidar_params}
A rotating \lidar contains several lasers stacked vertically. Each laser is given an index $r \in \{1, 2, \cdots N\}$, where $N$ is the number of lasers.
Each laser has an inclination angle $\xi_i$ from the horizontal, positive below the horizontal, the set of which is represented by $\xi$.
The lowest laser below the horizontal is assigned index $r = 1$ and $r = N$ index represents the highest laser, usually inclined above the horizontal.
The \lidar is mounted on the vehicle at a height $H_L$ above the wheelbase.
For rotating \lidar, the horizontal step angle $\Uppsi$ is uniform and can be calculated as $360^o/SamplesPerRotation$.
The sensor is considered the origin point for the coordinate system.
At each rotation, a new column of N samples is recorded and indexed with $c$.
The sensor returns a range image, a 2D depth image of range values $R_{r,c}~\forall~r\in \{1, 2, \cdots N\},~c\in \{1, 2, \cdots \Uppsi\}$.
While we assume the more prolific rotating \lidar, a solid state \lidar~\cite{roriz2021automotive} would have similar parameters.
Table~\ref{tab:symbol_summary} and Figure~\ref{fig:alpha_proof}, respectively, summarize and represent some of these parameters, using example values from real-world dataset~\cite{sun2020scalability}.

\subsection{Constraints}
\label{sec:constraints}
We assume the following constraints for the  validity of the \textit{detectability} model in Section~\ref{sec:model}:

\texttt{\textbf{C1}:} All \lidar beams encountering solid obstacles, including ground, within the max range of the \lidar, return with strong enough intensity from the first obstacle they encounter so that the return is recorded. This assumption holds in real-world except when;
\ca there are physical impediments in the air like dust, smoke, fog, or rain; or,
\cb adversarial objects with extremely reflective, absorbent, or transparent surfaces.
This constraint is partially alleviated by accommodating for \lidar beam energy attenuation resulting in reduced effective \lidar range, due to air impediments, as described in Section~\ref{sec:lidar_attenuate}.

\texttt{\textbf{C2}:} The obstacle's width, projected on a plane perpendicular to the \lidar beams falling on the obstacle, must be enough to have \lidar points returned from it.
For example, a thin sheet aligned parallel to the \lidar beam's direction can never be detected. According to the formula for circle arc length, a conservative bound can be established:
\begin{align}
    \text{Width Projected} &\geq \frac{\Uppsi D \pi}{180^o}.
\end{align}
Since $\Uppsi$ is small, the arc length is approximately equal, though always greater than the chord length for the same points.
Being a small value, this width is not a prohibitive constraint, \eg using values from Table~\ref{tab:symbol_summary} for Waymo Open Dataset~\cite{sun2020scalability}, the width constraint at max \lidar range of $75$ $m$ is just $17$ $cm$.

\texttt{\textbf{C3}:} At least one \lidar  point exists on the ground before the obstacle. This is an algorithmic constraint~\cite{bogoslavskyi17pfg}.
While we focus on the primary Top \lidar only in the rest of this work, additional low height limited field of view (FOV) \lidar sensors can be used for obstacles closer than $D_{min}$, \eg Front, Rear, Side-Left and, Side-Right \lidar in Waymo Open Dataset~\cite{sun2020scalability}.

Furthermore, the following are assumed initially ($\S$\ref{sub:model_simple}) and relaxed in later sections:

\texttt{\textbf{A1}:} Obstacle touches the ground at $90^o$, \ie  $\alpha_o = 90^o$. Relaxed in Sections~\ref{sub:model_incline} and \ref{sub:model_gap}.

\texttt{\textbf{A2}:} There is no ground inclination change between the vehicle and the obstacle. Relaxed in Section~\ref{sub:model_inclined_ground}.

\texttt{\textbf{A3}:} No noise in detected range. Relaxed in Section~\ref{sub:model_noise}.

\begin{table}[t]
    \centering
    \caption{\label{tab:symbol_summary}
        Symbols Summary}
    \begin{tabular}{c | l | rl}
        \toprule
        Symbols & Description & \multicolumn{2}{c}{Example Values\tnote{$^\ddagger$}}                       \\ \midrule

        $N$        		            & Count of lasers in a vertical array
            & $64$~\cite{sun2020scalability}, & $32$~\cite{lgsvl}                                           \\
        $\xi$      		            & \lidar Beam Angles set                           
            & $\{-2.4^o, \cdots, 17.6^o\}$~\cite{sun2020scalability}, & $\{-10.7^o, \cdots, 30.7^o\}$~\cite{lgsvl}    \\
        $H_L$      		            & Height of \lidar sensor                          
            & $2.184~m$~\cite{sun2020scalability}, & $2.312~m$~\cite{lgsvl}                                 \\
        $\Uppsi$   		            & \lidar horizontal angle step                     
            & $0.136^o$~\cite{sun2020scalability}, & $1^o$~\cite{lgsvl}                                     \\
        $D_{min}^{\lidar}$  		& Distance to first \lidar return                  
            & $6.7~m$~\cite{sun2020scalability}, & $3.9~m$~\cite{lgsvl}                                     \\
        $R_{max,clear}^{\lidar}$    & Distance range of the sensor in clear conditions                     
            & $75~m$~\cite{sun2020scalability}, & $100~m$~\cite{lgsvl}                                      \\
        $R_{max}$                   & Distance range of the sensor in current conditions
            & &                                                                                             \\
        $\alpha_o$ 		            & Inclination of obstacle from ground
            & \multicolumn{2}{c}{$\{0^o, \cdots, 8^o\}$~\cite{aashto2001policy}}                                 \\
        $h_o$      		            & Height of obstacle
            & \multicolumn{2}{c}{Variable}                                                                  \\
        $\alpha_{th}$ 	            & Threshold angle for ground removal
            & \multicolumn{2}{c}{$10^o$}                                                                    \\ \bottomrule
    \end{tabular}
    \begin{tablenotes}
        \footnotesize \centering
        \item[$^\ddagger$] Where different, values are provided for both evaluation setups ($\S$\ref{sec:eval_vod}, $\S$\ref{sec:eval_ps}).
    \end{tablenotes} 
\end{table}

\subsubsection{\lidar Beam Energy Attenuation}
\label{sec:lidar_attenuate}
Particulate impediments in air, \eg smoke, dust, or fog, reflect and dissipate \lidar beam energy cause false early or lost \lidar beam returns~\cite{bijelic2018benchmark,heinzler2019weather,wallace2020full,li2020happens}.
The detection of these conditions is an orthogonal problem with significant prior solutions~\cite{pavlic2012image,sallis2014air,straub2020detecting,az_dust,miclea2021visibility}.
Note that we only consider benign faults, not \lidar spoofing attacks~\cite{petit2015remote,shin2017illusion,cao2019adversarial,sun2020towards}, which can also violate constraint \texttt{\textbf{C1}}.
We now consider the reduction in the effective range of \lidar, from excessive attenuation of beam energy due to particulates in the air.
The power of \lidar attenuates per Beers-Lambert Law~\cite{weichel1990laser}:
\begin{align}
	\tau (R) = \frac{P(R)}{P(0)} = e^{-\sigma R}
	\label{eq:tau}
\end{align}
where $P(R)$, $P(0)$ refer \lidar power at distance $R$ and $0$, respectively. The value $\sigma$ is the attenuation coefficient that characterizes how strongly a medium absolves the wave signals.
The attenuation coefficient is calculated by the Kruse model~\cite{kaasalainen2008radiometric}
\begin{align}
	\sigma(\lambda) = \frac{17.35}{V}(\frac{\lambda}{0.55})^{-q}	
\end{align}
where $V$ is the visibility which depends on air quality and conditions, such as fog and haze. The parameter $\lambda$ is the wavelength of the signals in the micrometer. A typical \lidar wavelength is either 905 nm or 1550 nm.
The parameter $q$ depends on visibility~\cite{kim2001comparison} as
\begin{align}
	q &=
	\left\{
	\begin{array}{cc}
		1.6&V>50 \ km\\
		1.3&6 \ km <V<50 \ km\\
		0.16V+0.34 & 1 \ km <V<6 \ km\\
		V-0.5&0.5\ km <V<1 \ km\\
		0   &V<0.5 \ km\\
	\end{array}
	\right.    
\end{align}
Knowing the \lidar's visibility under the current weather condition is critical because the short-sighted perception will delay the vehicle's response and limit the ability to respond to obstacles. We emphasize this limitation but leave the accurate calculation of the measurable range as future work because this is out of the current paper's scope and has been studied in prior works~\cite{miclea2020laser}. Instead, we provide a rule of thumb.

Regardless of the weather condition, minimum observable signal power remains unchanged for a \lidar device, \ie there exists $\tau_{min}$ in~\eqref{eq:tau}. Therefore, we have
\begin{align}
    \tau_{min} = e^{-\sigma_{clear} R_{max,clear}^{\lidar}} = e^{-\sigma_{current} R_{max,current}^{\lidar}}
\end{align}
for clear weather and current weather.
By comparing the exponent, we have the maximum measurable distance of \lidar under the current weather as follows:
\begin{align}
    R_{max}^{\lidar} = \frac{\sigma_{clear}}{\sigma_{current} }R_{max,clear}^{\lidar}.
    \label{eq:r0}
\end{align}
    By definition, \mbox{$\sigma_{clear} \leq \sigma_{current}$}.
    Therefore, with the above observation, for current weather conditions, we get
\begin{align}
    R_{max}^{\lidar} \leq R_{max,clear}^{\lidar}. \label{eq:range_weather}
\end{align}	
Given the typical attenuation coefficients
  $0.1$ for clear air, $1$ for haze, and $10$ for fog~\cite{kim2001comparison},
  the effective range reduces $10\times$ in haze and $100\times$ in dense fog.
Going forward $R_{max}^{\lidar}$, shortened to $R_{max}$, is used to represent the effective range of the \lidar.
\section{Detectability Model}
\label{sec:model}

A \textit{detectability} model describes, given an obstacle's properties, sensor parameters, and environment parameters, whether a given algorithm can detect the obstacle with a given set of algorithm parameters.
The \textit{detectability} model allows the conversion of safety standards into an algorithm and sensor parameter requirements as described with the example in Section~\ref{sec:motivation}.
We now develop the \textit{detectability} model for obstacle detection using a \lidar.
We use Depth Clustering~\cite{bogoslavskyi16iros, bogoslavskyi17pfg} as the example algorithm, which uses depth discontinuity to segment \lidar points into ground \vs obstacles and then determine bounding boxes for obstacles.
Table~\ref{tab:symbol_summary} contains a summary of the symbols used in this section with example values from Waymo Open Dataset~\cite{sun2020scalability} or chosen defaults.

\begin{figure*}[t]
    \centering
    \begin{minipage}[t]{0.48\textwidth}
        \centering
        \includegraphics[width=.8\linewidth,keepaspectratio]{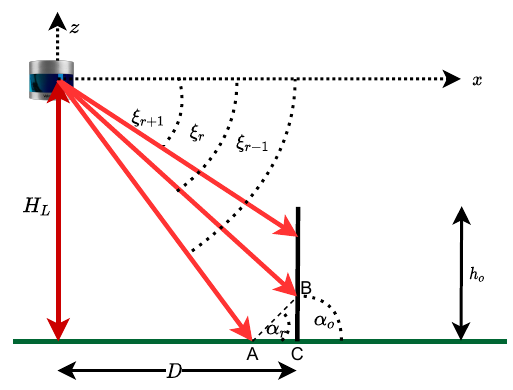}
        \caption{\label{fig:alpha_proof}
            A representation of point segmentation to ground \vs obstacle based on $\alpha$ thresholding when two points return from the obstacle.
        }
    \end{minipage}
    \hfill
    \begin{minipage}[t]{0.48\textwidth}
        \centering
        \includegraphics[width=.8\linewidth,keepaspectratio]{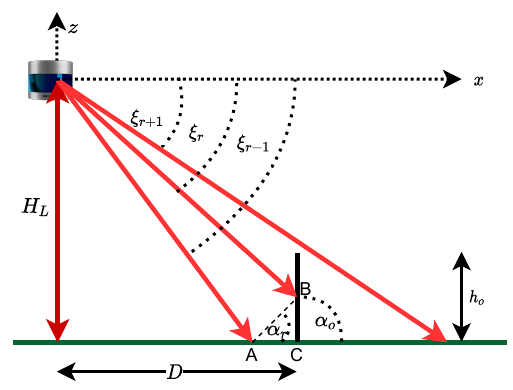}
        \caption{\label{fig:alpha_proof_one_point}
            Representation for $\alpha$ thresholding with only one point on the obstacle.
        }
    \end{minipage}
\end{figure*}

\subsection{Simple Model}
\label{sub:model_simple}

We start with a simple system model, as shown in Figures~\ref{fig:alpha_proof}~and~\ref{fig:alpha_proof_one_point}, with Assumptions \texttt{A1}, \texttt{A2} and \texttt{A3}.

\textit{Ground Removal:}
The primary part of the Depth Clustering algorithm that we consider in this work is ground removal, \ie determining which range image points are on obstacles \vs ground.
FN errors can occur if points on obstacles are mistakenly considered part of the ground.
Ground removal is based on vertical depth discontinuity using the inclination angles $\alpha$.
Assuming the first point ($r=1$) to always be on the ground, a sharp change in $\alpha$ above a threshold  $\alpha_{th}$ indicates that a point is on an obstacle.
The first point must lie on the ground for the subsequent comparisons to be meaningful (\texttt{C3}).
The inclination angle is calculated from the range image, as designed by Bogoslavskyi and Stachniss~\cite{bogoslavskyi17pfg}, shown in Figure~\ref{fig:alpha_proof}, as
\begin{align}
    \alpha_{r,c} &= atan2(||BC||, ||AC||)
    = atan2(\Delta z, \Delta x)
    = atan2(|R_{r-1,c} sin \xi_{r-1} - R_{r,c} sin \xi_{r}|,
         |R_{r-1,c} cos \xi_{r-1} - R_{r,c} cos \xi_{r}|)  \label{eq:alpha} \\
    \Delta \alpha_{r,c} &= \left\{\begin{array}{lr}
        0^o,\text{if } r=1 \\%
        |\alpha_{r,c} - \alpha_{r-1,c}|,\text{otherwise}%
        \end{array}\right\} \label{eq:delta_alphs}
\end{align}

The point $r,c$ is considered to not be on the ground if, $R_{r-1,c}$ is on the ground, $\alpha_{th} < 45^o$, and
\begin{align}
    \Delta \alpha_{r,c} &>  \alpha_{th}, \label{eq:alpha_condition}
\end{align}

If $R_{r-1,c}$ is not on the ground then $R_{r,c}$, is also considered to not be on the ground.
The lowest row of points $R_{0,c}$ are assumed to be on the ground ($\S$\ref{sec:constraints} {\tt C3}).
The column index corresponds to the rotational position of the sensor.
For brevity, the column index is omitted for points in the same column only.

At horizontal distance $D$ from the sensor, the height of a \lidar beam $r$ can be calculated as:
\begin{equation}
    H_r(D, \xi_r, H_L) = H_L - D * tan(\xi_r).
    \label{eq:Hr1}
\end{equation}

For brevity, $H_r(D, \xi_r, H_L)$ is referred to as $H_r(D)$ from this point as $\xi_r$ and $H_L$ are constant once a \lidar sensor is chosen and placed on an AV. Let us define $r = \min \{i|H_i(D) > 0^o\}$, \ie $R_r$ is the first point on the obstacle.

With this background, we can determine the minimum height of an obstacle that can be detected when the obstacle is at a distance of $D$ from the sensor (Figure~\ref{fig:alpha_proof}). It is required to check three consecutive \lidar points to detect an obstacle, according to the ground removal algorithm presented in~\eqref{eq:alpha} through~\eqref{eq:alpha_condition}.

\begin{theorem}
\label{th:the1}
    The obstacle is detected at a distance $D$, if and only if one of the following conditions is true:
    \begin{enumerate}
        \item $H_{r}(D) \leq h_o < H_{r+1}(D)$  AND $\alpha_{th} < atan2(H_{r}(D), |D - \frac{H_L}{tan(\xi_{r-1})}|)$;
        \item $h_o \geq H_{r+1}(D)$.
    \end{enumerate}
\end{theorem}

\begin{proof}
\textit{Sufficiency:}
Consider case 2), \ie $h_o \geq H_{r+1}(D)$. Then the points $R_r$ and $R_{r+1}$ are on the obstacle, as shown in Figure~\ref{fig:alpha_proof}, and thus it holds that $\alpha_{r+1}=90^o$.
Since we have
\begin{align*}
    &\Delta \alpha_r = \alpha_r-0^o,&&
    \Delta \alpha_{r+1} = |90^o-\alpha_r|,
\end{align*}
where  $\alpha_{r-1}=0^o$ by the definition of index $r$, it follows that
\begin{align*}
    \max\{\Delta \alpha_r,\Delta \alpha_{r+1}\} \geq 45^o>\alpha_{th},
\end{align*}
which guarantees that the obstacle is detected at a distance $D$.

Now consider case 1), \ie $H_{r}(D) \leq h_o < H_{r+1}(D)$ as shown in Figure~\ref{fig:alpha_proof_one_point}.
By the assumption, there exists an index $r-1$ that touches the ground.
Since $h_o \geq H_{r}(D)$, the angle $\alpha_{r-1}$ can be found by \eqref{eq:alpha} as follows:
\begin{align}
    \alpha_{r} &= atan2(H_{r}(D), |R_{r-1} cos (\xi_{r-1}) - R_{r} cos (\xi_{r})|) \nonumber\\
    &= atan2(H_{r}(D), |D - \frac{H_L}{tan(\xi_{r-1})}|),
    \label{eq:simple_cond}
\end{align}
where
\begin{align*}
    &R_{r-1}sin (\xi_{r-1}) = H_L,  &&R_{r} cos (\xi_{r}) = D.
\end{align*}

Since $\alpha_{r-1}=0^o$, the above equation and the angle condition in case 1) imply $\alpha_{th} < \Delta \alpha_{r}$, \ie the obstacle is detected at distance $D$.

\textit{Necessity:}
If the obstacle is detected at a distance $D$, then at least one point on the obstacle is at distance $D$, \ie $R_r$ is on the obstacle. This implies $h_o \geq H_{r}(D)$.
Furthermore, if the obstacle is detected, then
\begin{align}
    \Delta \alpha_{r} = \alpha_r-\alpha_{r-1} = \alpha_r > \alpha_{th}.
    \label{eq:alpha_th_cond}
\end{align}
There are two cases: $\alpha_{r+1}=90^o$ and $\alpha_{r+1} \neq 90^o$. If $\alpha_{r+1}=90^o$, then
$R_{r+1}$ is on the obstacle, and thus we have $h_o \geq H_{r+1}(D)$. This implies case 2).
On the other hand, if $\alpha_{r+1} \neq 90^o$, then
the inequality in~\eqref{eq:alpha_th_cond} must hold, which renders the condition $\alpha_{th} \leq atan2(H_{r}(D), |D - \frac{H_L}{tan(\xi_{r-1})}|)$.
This implies case 1). This completes the proof.
\end{proof}

Theorem~\ref{th:the1} consists of two sets of conditions based on the obstacle's height. The second set indicates that if the height of the obstacle is sufficiently large, then there will be more than two \lidar points on the obstacle ($\alpha_{r+1}=90^o$). This allows us to detect the obstacle without further condition on the angle $\Delta \alpha_r$.
The first set is the case that, when the size of the obstacle is small, there is only one \lidar point on the obstacle, which requires us to have an additional condition for the angle $\Delta \alpha_r$.

The first condition set in Theorem~\ref{th:the1} implies that the minimum height to be detected at distance $D$ satisfies
\begin{align}
   &H_{r}(D) = h_o \nonumber\\
   &\alpha_{th} \leq atan2(h_o, |D - \frac{H_L}{tan(\xi_{r-1})}|)
   \label{eq:minimum_height},
\end{align}
which depends on the distance $D$ and threshold $\alpha_{th}$.

\subsection{Obstacle at Inclination}
\label{sub:model_incline}
\noindent
\textit{Assumptions:}
We maintain all assumptions from $\S$\ref{sub:model_simple} except part of \texttt{A1}, \ie that the obstacle surface inclination angle is $\alpha_o \neq 90^o$. We assume $\alpha_o > \alpha_{th}$, otherwise, the obstacle cannot be detected.

In this case, if $R_{i}$ is at the end of the inclined obstacle, then its height at a distance $D$ is found by
\begin{align*}
    H_{i}(D) = h_o sin (\alpha_o) + h_o cos (\alpha_o) tan (\xi_{i}).
\end{align*}
This height can be used to determine whether there are more than two \lidar points on the obstacle. Further notice that the angle $\alpha_r$ is found by reduced height $\frac{H_r(D)}{tan(\xi_r)/tan(\alpha_o)+1}$ and increased width $D+\frac{H_r(D)}{tan(\xi_r)+tan(\alpha_o)}-\frac{H_L}{tan(\xi_{r-1})}$. These observations induce the following Corollary from Theorem~\ref{th:the1}.
\begin{corollary}
    The obstacle is detected at a distance $D$, if and only if one of the following conditions is true:
    \begin{enumerate}
        \item $H_{r}(D) \leq h_o sin (\alpha_o) + h_o cos (\alpha_o) tan (\xi_{r+1}) < H_{r+1}(D)$ AND $\alpha_{th} < atan2(\frac{H_r(D)}{tan(\xi_r)/tan(\alpha_o)+1}, |D+\frac{H_r(D)}{tan(\xi_r)+tan(\alpha_o)}-\frac{H_L}{tan(\xi_{r-1})}|)$;
        \item $h_o sin (\alpha_o) + h_o cos (\alpha_o) tan (\xi_{r+1}) \geq H_{r+1}(D)$.
    \end{enumerate}
\label{cor1}
\end{corollary}

\begin{figure*}[t]
    \begin{minipage}[t]{0.48\textwidth}
        \centering
        \includegraphics[width=.8\linewidth,keepaspectratio]{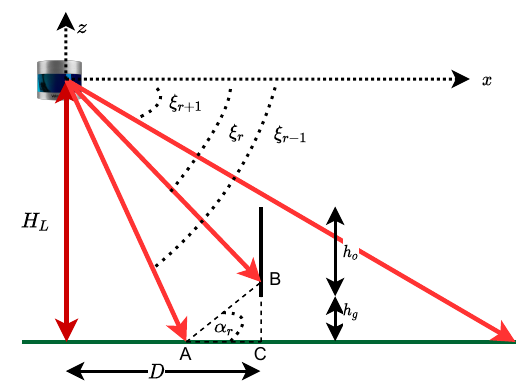}
        \caption{\label{fig:alpha_proof_gap}
            The obstacle elevated above the ground plane, causing a gap between the ground plane and the obstacle.
        }
    \end{minipage}
    \hfill
    \begin{minipage}[t]{0.48\textwidth}
        \centering
        \includegraphics[width=.8\linewidth,keepaspectratio]{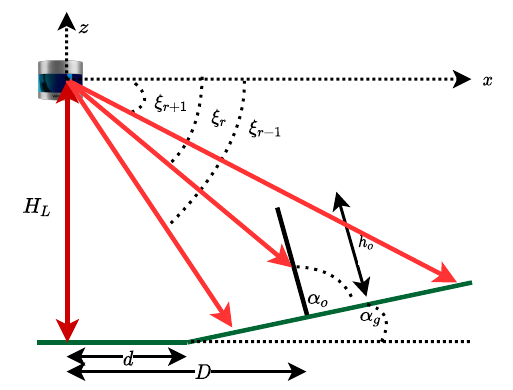}
        \caption{\label{fig:alpha_proof_ground_incline}
            Obstacle on the ground that is inclined relative to the AV. Note that only relative inclination change between AV and obstacle is material to the model.
        }
    \end{minipage}
\end{figure*}

\subsection{Obstacle not Touching Ground}
\label{sub:model_gap}

\noindent
\textit{Assumptions:}
We maintain all assumptions in Section~\ref{sub:model_simple} except \texttt{A1}, \ie that the obstacle surface no longer touches the ground. Instead, now the obstacle surface starts at height $h_g$ above the ground as shown in Figure~\ref{fig:alpha_proof_gap}.

We are interested in the first beam on the obstacle, where its index is defined by $r_g = \min \{i|H_i(D) > h_g\}$. Noticing the height of the obstacle tip is $h_o+h_g$, we can reformulate Theorem~\ref{th:the1} as the following corollary.
\begin{corollary}
    The obstacle is detected at a distance $D$, if and only if one of the following conditions is true:
    \begin{enumerate}
        \item $H_{r_g}(D) \leq h_o+h_g < H_{r_g+1}(D)$ AND $\alpha_{th} < atan2(H_{r_g}(D), |D - \frac{H_L}{tan(\xi_{r_g-1})}|)$;
        \item $h_o+h_g \geq H_{r_g+1}(D, \xi, H_L)$.
    \end{enumerate}
\label{cor2}
\end{corollary}

The minimum detectable height for this case is found by
\begin{align*}
   &H_{r_g}(D) = h_o + h_g \nonumber\\
   &\alpha_{th} \leq atan2(h_o + h_g, |D - \frac{H_L}{tan(\xi_{r_g-1})}|).
\end{align*}

\subsection{Inclined Ground}
\label{sub:model_inclined_ground}

\textit{Assumptions:}
We maintain all assumptions in Section~\ref{sub:model_simple} except \texttt{A2}, \ie Ground inclination $\alpha_g$ could be non-zero (Figure~\ref{fig:alpha_proof_ground_incline}).

Ground inclination affects the \textit{detectability} model only when relative inclination changes between the obstacle and the AV.
This section assumes that ground inclination starts before the obstacle position, \ie $d < D$.
There are two cases.

Case I: The inclination starts after the beam $R_{r-1}$, \ie
\begin{align}
	\frac{H_L}{tan(\xi_{r-1}) } \leq d \leq \frac{H_L}{tan(\xi_{r}) }.
\end{align}

We can extend Corollary~\ref{cor1} to find the \textit{detectability} condition. The first beam touching the obstacle must be higher than the ground at a distance of $D$. Let us define
\begin{align*}
    r_f = \min \{i|H_i(D) > (D-d) tan (\alpha_g)\}.
\end{align*}

If $R_{r_f+1}$ is at the end of the obstacle, then
\begin{align*}
    H_{r_f+1}(D) &= h_o cos (\alpha_g) - h_o sin (\alpha_g) tan (\xi_{r_f+1})
    + (D-d) tan (\alpha_g),
\end{align*}
which is the threshold to determine whether $R_{r_f+1}$ is on the obstacle.
Furthermore, $\alpha_{r_f}$ is found by increased height
$\frac{H_{r_f}(D)-(D-d) tan (\alpha_g)}{1-tan(\xi_{r_f})tan(\alpha_g)}+(D-d) tan (\alpha_g)$
and decreased width
$D+\frac{H_{r_f}(D)-(D-d) tan (\alpha_g)}{tan(\xi_{r_f})-cot(\alpha_g)}-\frac{H_L}{tan(\xi_{r_f-1})}$.
This observation induces the following:
\begin{corollary}
    The obstacle is detected at a distance $D$, if and only if one of the following conditions is true:
    \begin{enumerate}
        \item $H_{r_f}(D) \leq
        h_o cos (\alpha_g) - h_o sin (\alpha_g) tan (\xi_{r_f+1})+ (D-d)tan (\alpha_g)
        < H_{r_f+1}(D)$ AND $\alpha_{th} < atan2(\frac{H_{r_f}(D)-(D-d) tan (\alpha_g)}{1-tan(\xi_{r_f})tan(\alpha_g)}+(D-d) tan (\alpha_g), |D+\frac{H_{r_f}(D)-(D-d) tan (\alpha_g)}{tan(\xi_{r_f})-cot(\alpha_g)}-\frac{H_L}{tan(\xi_{r_f-1})}|)$;
        \item $h_o cos (\alpha_g) - h_o sin (\alpha_g) tan (\xi_{r_f+1})+ (D-d)tan (\alpha_g) \geq H_{r_f+1}(D)$.
    \end{enumerate}
\label{cor3}

\end{corollary}

Case II:
The inclination starts before the beam $R_{r-1}$, \ie
\begin{align*}
	d < \frac{H_L}{tan(\xi_{r-1}) }.
\end{align*}

In this case, the beam point of $R_{r_f-1}$ land on a different point from that of case II. This decreases relative height $H_L-R_{r_f-1} sin (\xi_{r_f-1})$ and increases relative width $\frac{H_L}{tan (\xi_{r_f-1})}-R_{r_f-1} cos (\xi_{r_f-1})$ between $R_{r_f-1}$ and $R_{r_f}$.
This changes $\alpha_{r_f}$ found in Corollary~\ref{cor3}.
It is also worth noticing that if
\begin{align*}
	d < \frac{H_L}{tan(\xi_{r-2}) },
\end{align*}
then $\alpha_{r_f-1} = \alpha_g$, and thus we have $\Delta \alpha_r = \alpha_r - \alpha_g$.
Using these facts, one can find an extension of Corollary~\ref{cor3}.
\subsection{Noise}
\label{sub:model_noise}

\textit{Assumptions:}
We maintain all assumptions in Section~\ref{sub:model_simple} except \texttt{A3}, \ie there exists a depth detection error.
Minor inaccuracies in the depth detection can be a problem when finding $\Delta x, \Delta z$.
In particular, the $i^{th}$ beam returns noisy measurement $y_i = R_i \epsilon_i$~\cite{wang2019pseudo} instead of its ground truth distance $R_i$ from the sensor to the obstacle where $\epsilon_i \in [1-\epsilon,1+\epsilon]$ is unknown noise with a known sensor error bound $\epsilon$, \eg Velodyne HDL-64E S3 \lidar  has a range detection inaccuracy of $\pm2$ $cm$.\footnote{\url{https://velodynelidar.com/products/hdl-64e/}}

Under mild assumptions, Theorems~\ref{th:the2} and~\ref{th:the3} provide a bound of the first angle $\alpha_r$ on the obstacle, and a bound of the angle $\alpha_{i}$ between two consecutive points on the obstacle.

\begin{theorem}
Assume
$y_{r}cos(\xi_r) \geq y_{r-1}cos(\xi_{r-1})$ and $y_{r}sin(\xi_r) \leq y_{r-1}sin(\xi_{r-1})$. Then, the angle $\alpha_r$ is lower bounded by
\begin{alignat*}{2}
\alpha_r &\geq atan2(&&R_{r-1}(1-\epsilon) sin(\xi_{r-1})-R_r(1+\epsilon) sin(\xi_r),
     R_r(1+\epsilon) cos(\xi_r)-R_{r-1}(1-\epsilon) cos(\xi_{r-1}))
\end{alignat*}
and upper bounded by
\begin{alignat*}{2}
\alpha_r &\leq atan2(&&R_{r-1}(1+\epsilon) sin(\xi_{r-1})-R_r(1-\epsilon) sin(\xi_r),
     R_r(1-\epsilon) cos(\xi_r)-R_{r-1}(1+\epsilon) cos(\xi_{r-1})).
\end{alignat*}
\label{th:the2}
\end{theorem}
\begin{proof}
Angle $\alpha_r$ is found by \eqref{eq:alpha}. The assumptions $y_{r}cos(\xi_r) \geq y_{r-1}cos(\xi_{r-1})$ and $y_{r}sin(\xi_r) \leq y_{r-1}sin(\xi_{r-1})$ imply that
\begin{alignat*}{2}
\alpha_r &= atan2(&&|y_r sin(\xi_r)-y_{r-1} sin(\xi_{r-1})|,
     |y_r cos(\xi_r)-y_{r-1} cos(\xi_{r-1})|)\nonumber\\
    &= atan2(&&y_{r-1} sin(\xi_{r-1})-y_r sin(\xi_r),
     y_r cos(\xi_r)-y_{r-1} cos(\xi_{r-1})).
\end{alignat*}
Considering the fact that $tan$ is a strictly increasing function in the domain $[0^o,90^o)$, we can find the lower bound and upper bounds presented in the theorem statement.
\end{proof}

\begin{theorem}
Assume that two consecutive points $R_{i-1}$ and $R_{i}$ land on the obstacle, and that
$y_{i}sin(\xi_i) \leq y_{i-1}sin(\xi_{i-1})$. Then, the angle $\alpha_i$ is upper bounded by $90^o$ and lower bounded by
\begin{alignat*}{2}
\alpha_i &\geq atan2(&&R_i(1+\epsilon) sin(\xi_i)-R_{i-1}(1-\epsilon) sin(\xi_{i-1}),
     R_i(1+\epsilon) cos(\xi_i)-R_{i-1}(1-\epsilon) cos(\xi_{i-1})).
\end{alignat*}
\label{th:the3}
\end{theorem}
\begin{proof}
The function $atan2$ is upper bounded by $90^o$ in all domains.
Similarly, the lower bound can be found with the proof of Theorem~\ref{th:the2}.
\end{proof}

The assumptions $y_{r}cos(\xi_r) \geq y_{r-1}cos(\xi_{r-1})$ and $y_{r}sin(\xi_r) \leq y_{r-1}sin(\xi_{r-1})$ are mild
because $R_{r}cos(\xi_r) \geq R_{r-1}cos(\xi_{r-1})$ and $R_{r}sin(\xi_r) \leq R_{r-1}sin(\xi_{r-1})$ hold for most of the cases, and $\epsilon_i$ is around $1$.
For the same reason, the assumption $y_{i}sin(\xi_i) \leq y_{i-1}sin(\xi_{i-1})$ is mild as well.

\subsection{Summary}
\label{sub:model_summary}

\newcommand{\maxfitplotsize}{.329}
\begin{figure*}[tp]
    \centering
    \subfloat[\centering \label{fig:hmin_simple} Simple Model ($\S$\ref{sub:model_simple})]{{\includegraphics[width=\maxfitplotsize\linewidth,keepaspectratio]{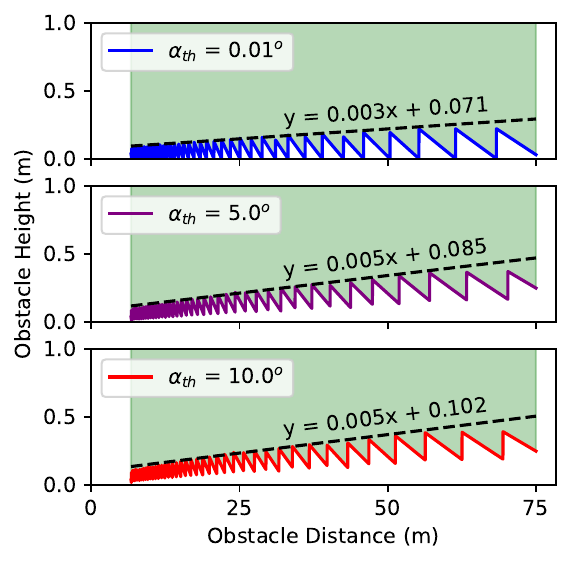}}}%
    \hfill
    \subfloat[\centering \label{fig:hmin_incline} $\alpha_o = 60^o$ ($\S$\ref{sub:model_incline})]{{\includegraphics[width=\maxfitplotsize\linewidth,keepaspectratio]{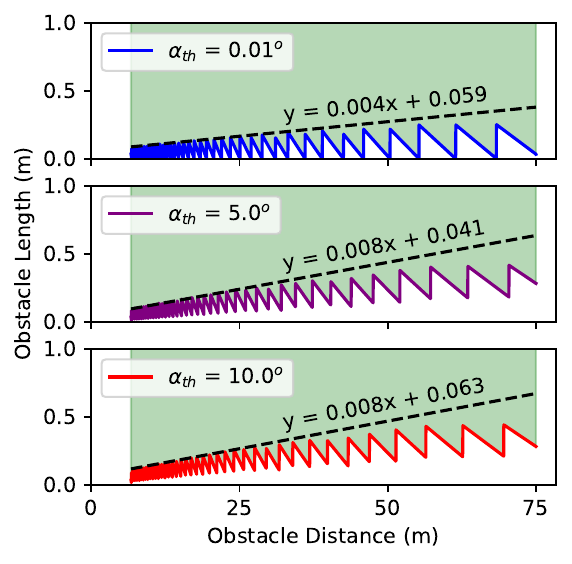}}}%
    \hfill
    \subfloat[\centering \label{fig:hmin_gap} $h_g = 0.271$ ($\S$\ref{sub:model_gap})]{{\includegraphics[width=\maxfitplotsize\linewidth,keepaspectratio]{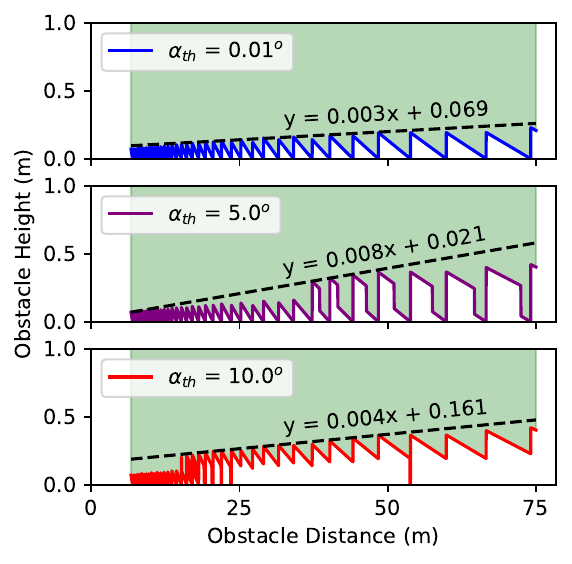}}}%
    \caption{
        Minimum detectable obstacle height (Y-Axis) at varying distances from the vehicle (X-Axis).
        Black dashed lines are a linear max fit for the plots, with corresponding equations.
        The shaded region signifies where the obstacle is always detected.}%
    \label{fig:hmin}%
\end{figure*}

Using $\xi$ and $H_L$ from the Waymo Open Dataset~\cite{sun2020scalability}, \eqref{eq:minimum_height} yields Figure~\ref{fig:hmin_simple} for various $\alpha_{th}$ and $D$ (X-Axis).
Figure~\ref{fig:hmin_incline} shows the same when $\alpha_o = 60^o$ as in Section~\ref{sub:model_incline}.
Figure~\ref{fig:hmin_gap} is based on Section~\ref{sub:model_gap}, assuming the obstacle is $0.271$ $m$ above the ground, based on the ground clearance of a sedan.
The varying nature of the colored plot is due to the discrete nature of \lidar beams.
To get a usable bound, like \eqref{eq:ideal}, we determine a linear max curve fit for each case, where the dashed line is always greater than or equal to the original plot. The shaded region is where the obstacle is always detected.

In this section, we have developed a \textit{detectability} model for the ground removal of the Depth Clustering algorithm.
We find that not only is such analysis possible, but it also yields human perceptible bounds.
This work provides bounds on the capabilities of the algorithm.
This work stands as the first step in establishing the verifiability of classical obstacle detection algorithms and their practical use in AV for safety-critical obstacle avoidance.
A limitation of the presented model is that while it describes the behavior of the core ground vs obstacle differentiation functionality,
  it does not cover features that reduce false positives or draw boundaries on individual obstacles.
We now show present the \ps design that integrates this algorithm into the perception pipeline to provide guarantees for collision avoidance.

\section{\theterm}
\label{sec:ps}

\begin{figure}[t]
	\centering
	\includegraphics[width=.65\linewidth]{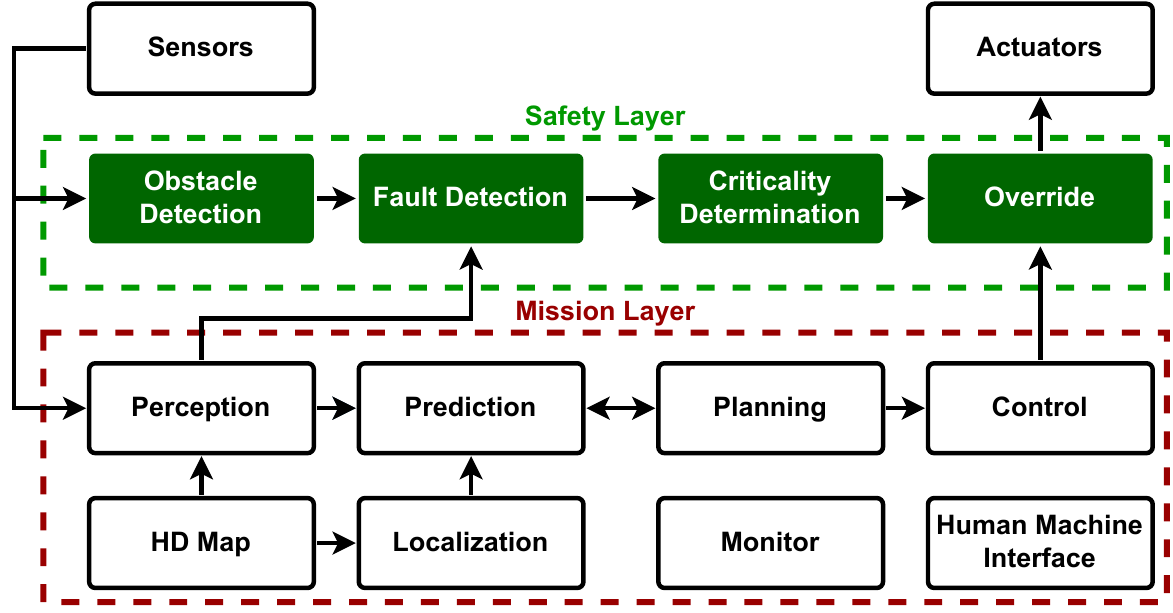}
	\caption{\label{fig:pipeline}
		Control and Data Pipeline of an Autonomous Driving system within \theterm (\ps) architecture.
		The mission layer uses Baidu's Apollo 5.0~\cite{apollo}.
		The safety layer uses Depth Clustering~\cite{bogoslavskyi17pfg} for obstacle detection.
	}
\end{figure}

\theterm (\ps) presents a system architecture that can utilize the verifiable obstacle detection capabilities described in Section~\ref{sec:model}.
Acknowledging the limitations on capabilities of verifiable algorithms and unverifiability of DNN,
  the goal of \ps is to provide specific safety guarantees amidst perception faults,
  rather than complete replacement as in traditional Simplex architecture~\cite{simplex_original,simplex2}.
As noted before in Section~\ref{sec:motivation}, this work focuses on obstacle existence detection faults.
Therefore \ps preserves the full functionality of the high-capability mission-critical software when no faults are detected.
When a fault is detected, and such a fault poses a risk of collision, the safety layer intervenes, abandoning the mission of the AV, to ensure safety.
Figure~\ref{fig:pipeline} shows the AV pipeline under \ps.

\subsection{Mission Layer}
\label{sec:spec_od}
\label{sec:impl_mission}

    The mission layer is an autonomous driving agent capable of fulfilling the mission of the AV, \ie driving from one point to another while trying to maintain safety.
However, if the safety layer detects a reason to override the mission layer, the mission of the AV is abandoned till the mission layer faults are mitigated.
The clear separation of the mission layer from the safety layer makes the \ps architecture forward compatible with all improvements in the autonomous driving agents.
The only explicit design requirement for the mission layer in \ps is to explore its intermediate results and control commands to the safety layer.
This is easily achieved in the publisher-subscriber model, common to the middleware of open-source AV implementations, \ie
  CyberRT~\cite{cyber-rt} for Baidu's Apollo 5.0~\cite{apollo} or ROS~\cite{ros} for Autoware~\cite{autoware}.
In this work, Baidu's Apollo 5.0~\cite{apollo} is used as the mission layer.

\subsection{Safety Layer - Obstacle Detection}
\label{sec:impl_alg}
Obstacle detection is performed using the Depth Clustering~\cite{bogoslavskyi16iros,bogoslavskyi17pfg} algorithm.
Figure~\ref{fig:det_eq} shows the \textit{detectability} model based on Corollary~\ref{cor2} and the parameters from evaluation setup~($\S$\ref{sec:eval_vod}).
It should be noted that while the \textit{detectability} model covers the core ground removal,
  as in \eqref{eq:alpha}, \eqref{eq:delta_alphs}, and \eqref{eq:alpha_condition},
  Depth Clustering algorithm has broader features, \eg points clustering to determine obstacle boundaries or noise filtering,
  that are used for the safety layer obstacle detection.

\subsection{Safety Layer - Fault Detection}
\label{sec:fr2}
Faults in mission layer object detection output are determined by comparing it against the safety layer's obstacle detection output.
For the comparison, we use the minimized sufficient obstacle detection requirements~($\S$\ref{sec:minimal_ca}).
Intuitively, the comparison checks whether mission layer detections cover the hazards detected by the safety layer.

\subsection{Safety Layer - Criticality Determination}
\label{sec:fr3}
\begin{figure}[t]
	\begin{minipage}[t]{0.48\textwidth}
		\centering
		\includegraphics[width=.9\linewidth,keepaspectratio]{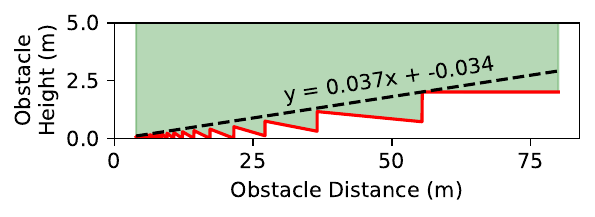}
		\caption[something]{\label{fig:det_eq}
		\textit{Detectability} model for the software-in-the-loop simualation setup ($\S$\ref{sec:eval_ps}).
			The red line plot uses Corollary~\ref{cor2} and parameters from the simulation setup.
			The black line is the linear max fit, same as Figure~\ref{fig:hmin}.
		}
	\end{minipage}
	\hfill
	\begin{minipage}[t]{0.48\textwidth}
		\centering
		\includegraphics[width=.7\linewidth,keepaspectratio]{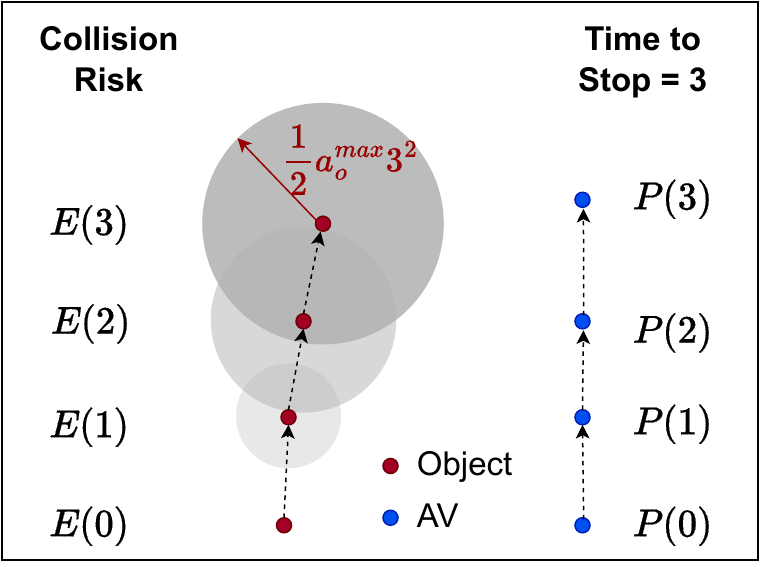}
		\caption[something]{\label{fig:risk}
			Collision risk within Time to Stop of the AV, adapted from prior work~\cite{9460196}.
			$E(t)$ denotes the existence region~\cite{schmidt2006research} of the obstacle at time $t$,
				\ie the region it can exist based on current position and velocity measurement,
				and maximum acceleration capability of $a_o^{max}$.
			$P(t)$ denotes the position of the AV on the local planned path at time $t$.
		}
	\end{minipage}
\end{figure}

To determine the appropriate reaction to a fault, first, its severity must be established.
For any obstacle existence detection fault, it must be ascertained if this obstacle poses a risk of collision with the AV.
We use a physical model for collision risk~\cite{9460196},
wherein an obstacle is determined to pose a risk of collision if their existence regions~\cite{schmidt2006research} overlap within the AV's time to stop.
Figure~\ref{fig:risk} describes the collision risk determination.

\subsection{Safety Layer - Override}
Two overrides are imposed on the mission layer control output.
Velocity is limited to maintain the safety guarantee, as derived later in Theorem~\ref{lem:safe} and \eqref{eq:vsafe_clear}.
An emergency brake is applied to avoid the potential collision if a safety-critical obstacle existence detection fault is determined and evaluated using Algorithm~\ref{alg:fault_alg}.

\begin{algorithm}[t]
  \DontPrintSemicolon
  \SetKwFunction{Match}{isDetected}
  \SetKwFunction{Risk}{isCollisionRisk}
  \caption{\label{alg:fault_alg}
	  Decision Logic.
  }

  \KwResult{No Override or Brake}

  \KwIn{
    $O_m$ : Mission layer's current set of objects,~ 
    $O_s$ : Safety layer's current set of obstacles
  }

  \BlankLine

  \If(\hskip 8mm \tcp*[h]{Run when either detection set changes}){$O_m.update()~||~O_s.update()$}
  {
    \ForEach{$o_s$ in $O_s$}{
      \If{\Match{$o_s, O_m$}}{
        continue
      }
      \If{\Risk{$o_s$}}{
        \KwRet (Brake)
      }
    }
  }
  \KwRet (No Override)

  \BlankLine

  \SetKwProg{Fn}{Function}{\string:}{}
  \Fn{\Match{$o_s, O_m$}}
  {
    Safety-critical minimized obstacle detection requirements ($\S$\ref{sec:minimal_ca}, $\S$\ref{sec:fr2}).
  }

  \BlankLine

  \SetKwProg{Fn}{Function}{\string:}{}
  \Fn{\Risk{$o_s$}}
  {
  	Potential for Collision~\cite{9460196}
  	using \mbox{Existence Regions}~\cite{schmidt2006research} ($\S$\ref{sec:fr3}, Figure~\ref{fig:risk}).
  }

\end{algorithm}

\subsection{Safety Guarantee}
Algorithm~\ref{alg:fault_alg} describes the safety layer decision logic.
We can now derive \ps's safety guarantee.

\begin{theorem}
	Let's assume an AV must detect all obstacles with height $\geq H^O$, defined by safety regulations.
	When \textit{detectability} model constraints are met, collisions with stationary obstacles of height $\geq H^O$, and that the mission layer fails to detect, are always avoided, when the AV's operating velocity is always below a predetermined limit:
	\begin{align}
		v^{safe}_{max} = \sqrt{(a^{av}_{max}L_{max})^2+2 a^{av}_{max}D_{max}^{stop}} -a^{av}_{max}L_{max}  \label{eq:vsafe_clear}
	\end{align}
	where $a^{av}_{max}$, $L_{max}$ are the maximum safe deceleration of the AV, and 
	worst-case computational latency, respectively. Furthermore, maximum stopping distance $D_{max}^{stop}$ depends on the \textit{detectability} model and the current maximum detection range of \lidar.
	\label{lem:safe}
\end{theorem}

\begin{proof}
	We will first show that $v^{safe}_{max}$ in~\eqref{eq:vsafe_clear} is the maximum safe velocity for collision avoidance by the safety layer.
	Let's denote the maximum range of the \lidar sensor in clear weather, a sensor specification, by $R_{max, clear}^{\lidar}$.
	This is a sensor specification, \eg $75~m$ for top \lidar in Waymo Open Dataset~\cite{sun2020scalability} and $100~m$ for the \lidar used in simulation~\cite{lgsvl_lidar}.
	As described in Section~\ref{sec:lidar_attenuate}, this range reduces in low visibility conditions, \eg due to fog or haze.
	The maximum detection range of \lidar $R_{max}^{\lidar}$ is as in \eqref{eq:r0}. 
	Using the \textit{detectability} model ($\S$\ref{sec:model}) the maximum range at which an object of height $H^O$ can be detected can be determined.
	Let's denote this range as $R^O_{max}$. 
	The maximum range at which obstacles can be reliably detected ($R_{max}$) is then
	\begin{align}
		R_{max} &= min(R^O_{max}, R_{max}^{\lidar}).\label{eq:det_max}
	\end{align}
	\eqref{eq:det_max} establishes the maximum detection range.
	However, as a safety policy and to accommodate for any error allowances within the system, a safety margin can be added to this range, denoted as $D^{Safety}$.
	Combining this with \eqref{eq:det_max} we get the maximum stopping distance for the AV
	and can now derive $v^{safe}_{max}$ by solving:
	\begin{align}
		D_{max}^{stop} &= R_{max} - D^{Safety} = v^{safe}_{max} (t+L_{max}) - \frac{1}{2} a^{av}_{max} t^2 \label{eq:vsolve_1}  
	\end{align}
	where $t$ is deceleration time. Substituting $t$ in \eqref{eq:vsolve_1} with $t= \frac{v^{safe}_{max}}{a^{av}_{max}}$,
	we get 
	\eqref{eq:vsafe_clear}.

	Now we prove the rest of the statement.
	Since the obstacle height is $\geq H^O$, the obstacle's existence is detected by the safety layer at all distances $\leq R_{max}$ in~\eqref{eq:det_max} \ie it is part of the set of obstacles detected by safety layer ($O_s$).
	When obstacle $o_s$ is not detected by the mission layer, \ie it is not part of the set of mission layer detections ($O_m$)
	\ie $o_s \notin O_m$, but $o_s \in O_s$, Algorithm~\ref{alg:fault_alg} returns "Brake" if this obstacle poses collision risk.
	If the current velocity is less than $v^{safe}_{max}$ in~\eqref{eq:vsafe_clear}, we can guarantee collision avoidance for the obstacle $o_s$. 
\end{proof}

It should be noted that while Theorem~\ref{lem:safe} considers stationary obstacles, trivially, any obstacle moving away from the AV's position is also avoided.
Even when AV velocity is higher than $v^{safe}_{max}$, braking will reduce collision velocity resulting in a reduced impact of any collisions.
Further, since all parameters here can be determined statically or based on AV's operating conditions (\eg $a^{av}_{max}$), $v^{safe}_{max}$ can be calculated to provide a guarantee against collisions with static obstacles.
Only $H^O$ is based on obstacle properties, which we assume are lower bound by regulations and policies ($\S$\ref{sec:motivation}).
There needs to be a policy or safety standard defining the minimum height of obstacles that must be detected and avoided by AV.
Given such a policy the operational design domain of the presented \ps design is determined by
  \ca adherence to the constraints of the \textit{detectability} model ($\S$\ref{sec:paramconstraints});
  \cb $v^{safe}_{max}$, adjustable by selecting sensor, AV and algorithm parameters.

\section{Dataset Evaluation}
\label{sec:eval_vod}

\begin{table}[tp]
    \centering
    \caption{\label{tab:eval_clear}
        Results for False Negative (FN) evaluation using real-world dataset.}
    \begin{tabular}{r|l}
        \toprule
        Count       & Category                                                                                  \\
        \midrule
        98224    	& Total obstacles (Vehicles, Pedestrian, Cyclist, Unknown)                                  \\
        93030    	& Total without obstacles closer than~$D_{min}$~(\texttt{C3}~$\S$\ref{sec:constraints})    \\
        7565     	& Obstacles that pose collision risk~($\S$\ref{sub:rrr})~\cite{9460196}                     \\
        7402		& Detected meeting minimally sufficient requirements~($\S$\ref{sec:minimal_ca})             \\
        153    		& Ground Truth larger than obstacle                                                         \\
        10    		& Oversegmentation                                                                          \\
        0		    & Remaining FN count at 75\% coverage                                                       \\
        \bottomrule
    \end{tabular}
\end{table}

In this section, we evaluate the obstacle detection algorithm using real-world sensor data from Waymo Open Dataset~\cite{sun2020scalability}.
We first determine if the algorithm meets the safety-critical requirements proposed in this work ($\S$\ref{sec:minimal_ca}).
For obstacle existence fault or FN evaluation, we randomly select 16 clear weather scenes from the 202 scenes in the validation dataset.
The random downsampling was necessitated by the manual effort involved in analyzing FN candidates.
The scenes included various scenarios, including heavy to low traffic, residential with pedestrians, city, and highway driving.

Table~\ref{tab:symbol_summary} summarizes the parameters used. $\alpha_{th}$ was set at $10^o$.
Distance overestimation error was bounded to $10$ $cm + 5\%$ of actual distance. The constant was chosen as a small value to keep the error bound low at close distances.
However, since the depth perception accuracy reduces with distance an additional percentage-based threshold allows accommodation of sensor limits while having a low impact.
A limited coverage threshold of $75\%$ was used as a threshold for True Positive (TP) detection, chosen to parallel the strict metric for IOU from COCO Dataset~\cite{coco}.
All FN candidates were analyzed manually using a visualizer provided by Bogoslavskyi and Stachniss~\cite{bogoslavskyi17pfg}.
We enhanced the visualizer with selective Ground Truth (GT) annotations and point cloud coloring, to aid the manual inspection.
Table~\ref{tab:eval_clear} summarizes the results.

\subsubsection{GT Counts}

We first determine the total number of GT in the included scenes. All classes, excluding road signs, were counted to yield a total of \textbf{98224}.
Obstacles closer than the first beam ($R_{1,c}$) on the ground, \ie closer than $D_{min}$ are removed as per \texttt{C3} in Section~\ref{sec:constraints}, reducing the count of GT to \textbf{93030}.
Obstacles that do not pose a risk of collision are also removed~($\S$\ref{sub:rrr}), leaving \textbf{7565} obstacles.

\subsubsection{Automated Evaluation}
We run the detection algorithm and evaluate results based on requirements described in Section~\ref{sec:minimal_ca}. \textbf{7402} True Positives and \textbf{163} FN candidates were found. The FN candidates were then manually inspected.

\subsubsection{GT Dimension Error}
\label{sec:gt_execption}

\begin{figure}[t]
    \centering
    \hfill
    \subfloat[\centering \label{fig:gt_too_large}Example showing 3D ground truth box extending too far beyond the obstacle.]
    {{\fbox{\centering\includegraphics[width=.533\linewidth,keepaspectratio]{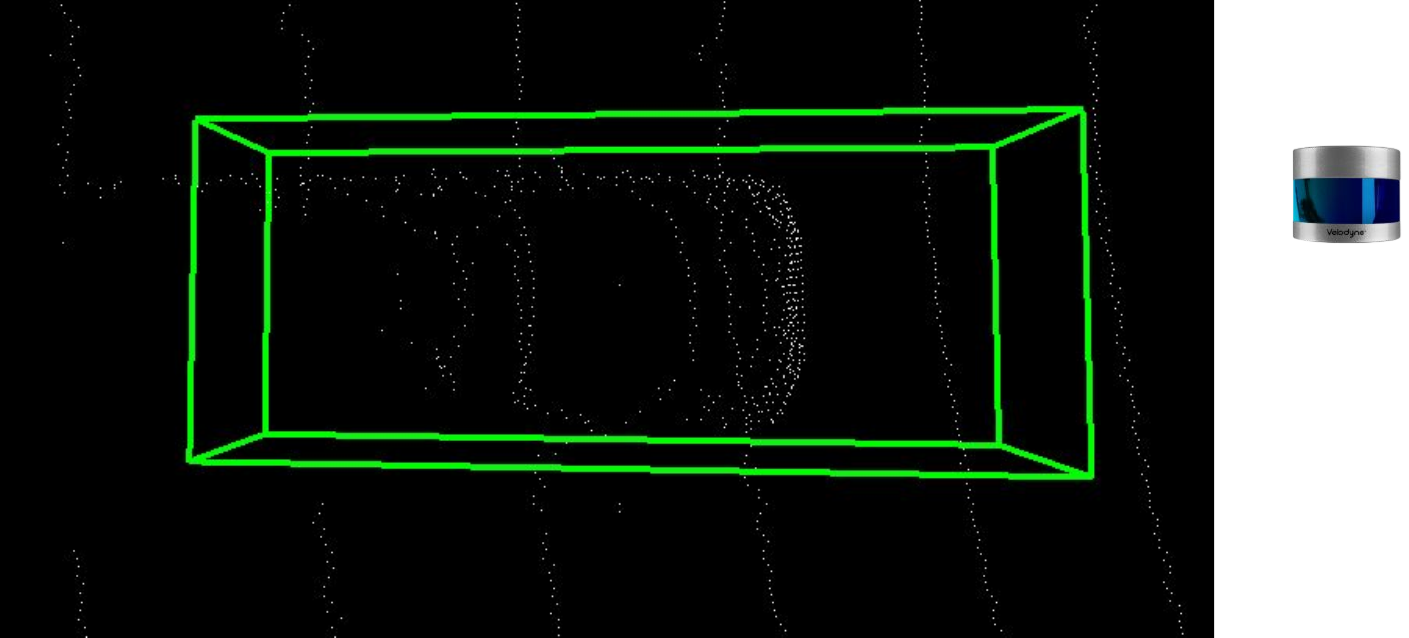}}}}%
    \hfill
    \subfloat[\centering \label{fig:coverage_example} GT extra size causing coverage (73.5\% here) to fall just below the threshold (75\%).]
    {{\fbox{\includegraphics[width=.35\linewidth,keepaspectratio]{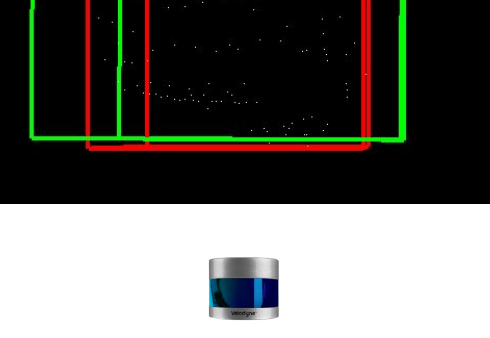}}}}%
    \hfill
    \\
    \hfill
    \subfloat[\centering \label{fig:oversegment} Example of Oversegmentation]
    {{\fbox{\includegraphics[width=.488\linewidth,keepaspectratio]{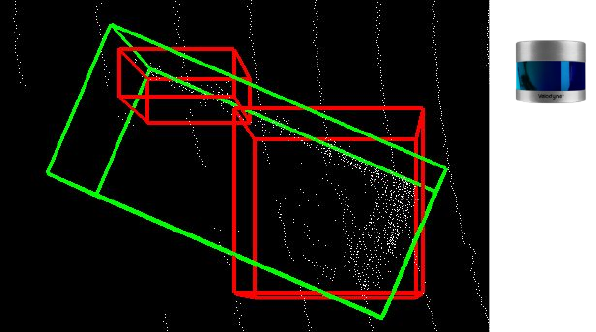}}}}%
    \hfill
    \subfloat[\centering \label{fig:dist_threshold} TP as the detection includes the visible edge of the vehicle towards the AV.]
    {{\fbox{\includegraphics[width=.35\linewidth,keepaspectratio]{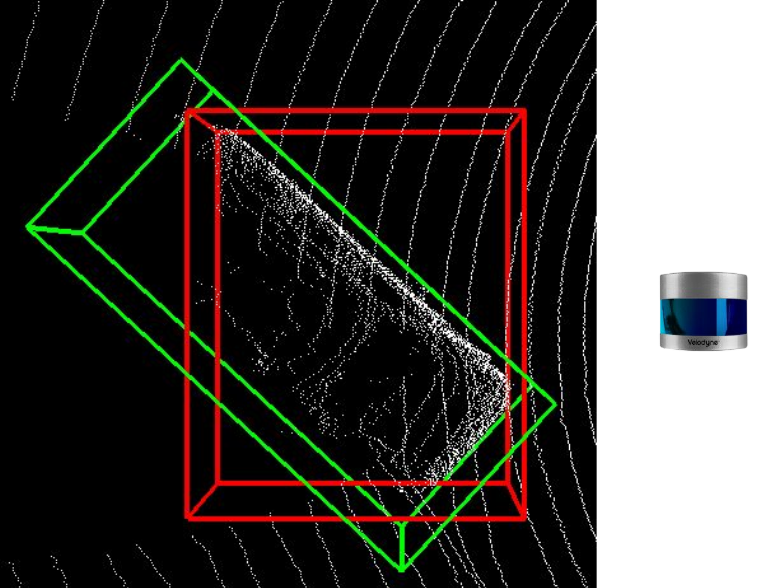}}}}%
    \hfill
    \caption{\label{fig:gt_issues}
        Examples from manual inspections of FN candidates.
        The white points are in 3D space as returned by \lidar \xspace \ie the point cloud.
        Ground truth 3D labels are drawn as green bounding boxes, while detection bounding boxes are red.
        The point cloud is shown from a top view, and the \lidar direction is indicated with the sensor image~\cite{puck_lidar}.
    }
\end{figure}

The most common reason for erroneous FN indication was inaccurate GT labels.
This inaccuracy of GT was determined based on a visual inspection of point clouds.
The GT was visibly larger or offset from the actual obstacle.
Figures~\ref{fig:gt_too_large}~and~\ref{fig:coverage_example} show such examples.
Drawn as per the provided GT, the green box is either larger or offset from the contained obstacle.
Similarly, as shown in Figure~\ref{fig:coverage_example}, the GT inaccuracy is enough to bring the coverage below the $75\%$ threshold.
In some cases, the error is small, \eg Figure~\ref{fig:dist_threshold}. However, when close to the AV, the small GT error can still be larger than the distance overestimation error allowed. We consider these detections as TP after ascertaining that the detection bounding box meets the requirements.
A total of \textbf{153} FN candidates were found to fall in this category.

\subsubsection{Oversegmentation}

Figure~\ref{fig:oversegment} shows a case where the obstacle was adequately detected but segmented into more than one bounding box. Since the second bounding box did not meet the distance threshold, the automated analysis ignored it.
However, given the presence of both bounding boxes, we argue that this detection should be considered True Positive.
The points on the obstacle were not erroneously considered to be drivable ground.
\textbf{10 }instances of this scenario on the same vehicle were found in consecutive frames.

\subsubsection{Obstacle Existence Fault}

No FN, \ie obstacle existence faults were found.
This is not surprising given the low minimum height bounds determined in Section~\ref{sec:model} and Figure~\ref{fig:hmin}.
The max curve fits in Figures~\ref{fig:hmin_simple},~\ref{fig:hmin_incline}~and~\ref{fig:hmin_gap} are conservative linear approximations.
Ignoring the max curve fits, the actual bounds in Figures~\ref{fig:hmin_simple},~\ref{fig:hmin_incline}~and~\ref{fig:hmin_gap} were less than $0.5$ $m$ in all cases within the sensor range of $75$ $m$.

\section{Software-in-the-loop Simulation}
\label{sec:eval_ps}

\begin{figure}[t]
	\centering
	\includegraphics[width=.6\linewidth]{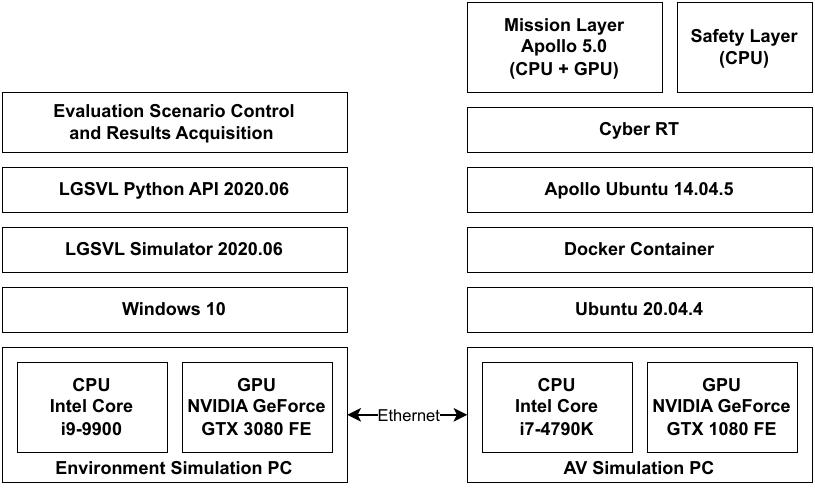}
	\caption{\label{fig:sim_platform}
		Software-in-the-loop simulation setup for the evaluation.
		LGSVL~\cite{lgsvl} simulates the driving environment, controlled via its Python API interface.
		The mission and safety layers components are shown in Figure~\ref{fig:pipeline}.
	}
\end{figure}

We evaluate the efficacy of \ps using software-in-the-loop simulation.
Figure~\ref{fig:sim_platform} describes the setup, based on Apollo 5.0~\cite{apollo} and LGSVL Simulator~\cite{lgsvl}.

\subsection{Setup}
\label{sec:eval_setup}

To evaluate the efficacy of \ps against obstacle existence faults, the scenario as shown in Figure~\ref{fig:estop_scenario} is used.
The scenario starts with the AV moving with initial velocity $v(0) \in \{ 5, 10, \cdots 40\}$ $m/s$
  with a vehicle is its path at distance, $d(0)  \in \{10, 20, \cdots 100\}$ $m$.
The AV's mission is to pass this obstacle.
The safety requirement during this mission is to avoid any collisions.
Each simulation run results in one of the following outcomes, summarized in Table~\ref{tab:outcomes}.
\begin{itemize}
	\item[{\color{red}\textbf{x}}]  \textit{Collision}: AV collides with the obstacle, therefore the safety requirement is violated and the mission cannot be fulfilled.
	\item[\tikzcircle{2pt}]  \textit{Safe Stop}: AV safely stops without colliding with the obstacle, however, it is unable to continue and pass the obstacle.
	\item[{\color{ForestGreen}\textbf{+}}] \textit{Safe Pass}: AV passes the obstacle without colliding with it.
\end{itemize}
After the simulation start ($t>0$),  \ps controls the AV in one of the following three configurations of the \ps pipeline  (Figure~\ref{fig:pipeline}):
\begin{itemize}
	\item \textit{Mission Only} ($\mathcal{MO}$): Mission Layer directly controls the AV.
		The safety layer is disabled.
		While this scenario provides the baseline for the mission capabilities of the system,
		however, the mission layer has considerations beyond collision avoidance, \eg mission completion and rider comfort, which makes a direct comparison less useful.
	\item \textit{Mission Crash} ($\mathcal{MC}$): Mission layer becomes unavailable at $t=0$,
	therefore brakes are applied $\forall t>0$ and hence is the synthetic ideal case for safety.
	\item \textit{Fault Injected} ($\mathcal{FI}$): All components of \ps are functional and available.
	However, obstacle existence detection faults are synthetically injected into the object detection output communicated from the mission layer to the safety layer.
	Safe handling of obstacle existence detection faults is the primary focus of this work ($\S$\ref{sec:scope}), therefore, this configuration represents the evaluation target.
\end{itemize}

$\mathcal{MO}$ setup provides a baseline for the mission behavior, while
$\mathcal{MC}$ shows the best-case braking response, given the dynamics of the scenario.
$\mathcal{FI}$ is the evaluation target.
Using simulation parameters,
$R^{\lidar}_{max,clear}=100$ $m$, $D^{Safety}=0.1$ $m$, $a^{av}_{max}=7.5$ $m/s^2$, $L_{max}=0.01$ $s$, and
$H_O = 0.75$ $m$ for sedan's back with \eqref{eq:vsafe_clear} we get $v^{safe}_{max} = 17.71$ $m/s$.
For the evaluation, the velocity limit is not enforced by the safety layer, rather it is used here to compare the \ps behavior when AV is below this limit \textit{vs} not.

\begin{table}[t]
	\hfill
	\begin{minipage}{0.5\linewidth}
		\centering
		\includegraphics[width=.9\linewidth,keepaspectratio]{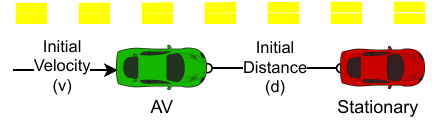}
		\captionof{figure}{
			\label{fig:estop_scenario}
			Evaluation scenario.
		}
	\end{minipage}
\hfill
	\begin{varwidth}{0.4\linewidth}
	\centering
	\caption{\label{tab:outcomes} Outcomes}
	\begin{tabular}{c@{\hskip 5mm} r@{\hskip 5mm} r}
		\toprule
		& \multicolumn{1}{c}{Safety} & \multicolumn{1}{c}{Mission}
		\\ \midrule
		Collision & Violated \xmark & Violated \xmark \\
		Safe Stop & Met \cmark & Violated \xmark \\
		Safe Pass & Met \cmark & Met \cmark 
		\\ \bottomrule
		%
	\end{tabular}%
\end{varwidth}%
\hfill
\end{table}
\subsection{Safety - Overview}
\label{sec:eval_outcomes}

\begin{figure}[t]
	\centering
	\subfloat[\label{fig:outcomes_below}$v(0) \leq v^{safe}_{max}$]{%
		\includegraphics[width=.54\columnwidth,keepaspectratio,trim={0 5mm 0 0}]{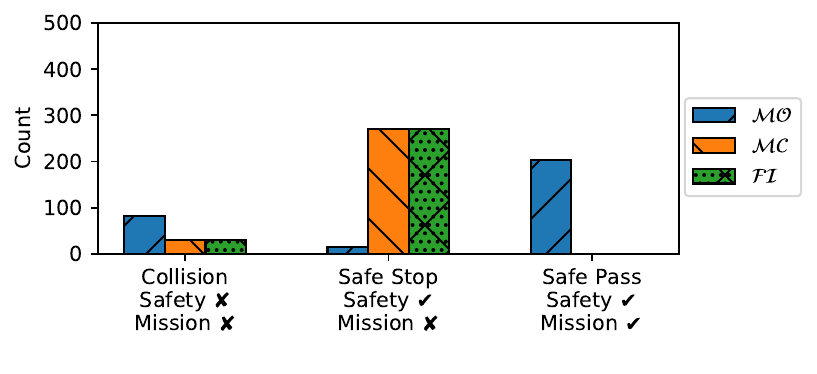}}
	\hfill
	\subfloat[\label{fig:outcomes_above}$v(0) > v^{safe}_{max}$]{%
		\includegraphics[width=.46\columnwidth,keepaspectratio,trim={0 5mm 0 0}]{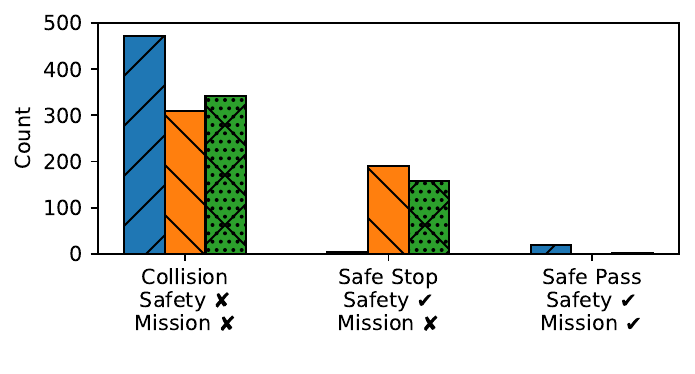}}
	\caption{\label{fig:outcomes}
		Comparison of scenario outcomes (X-Axis).
		Collision, Safe Stop, and Safe Pass are the possible outcomes (Table~\ref{tab:outcomes}).
		The Y-Axis counts how often an outcome is observed among all the simulation instances for each configuration ($\S$\ref{sec:eval_setup}).
	}
\end{figure}

For each $v(0), d(0)$ and \ps configuration combination, 10 simulation repetitions are run \ie a total of 2400 instances.%
\footnote{
	Out of any 10 repetitions, at minimum 8 instances yielded valid outcomes.
	Rarely ($0.3\%$), when initial velocity is high ($35~m/s~or~40~m/s$),
	AV becomes unstable and goes out of the road boundaries.
	Such instances are not included in the results.
}
We now compare the prevalence of different outcomes.
Figure~\ref{fig:outcomes_below} shows the count of outcomes when initial velocity is below the safe limit, \ie $v(0) \leq v^{safe}_{max}$ and
Figure~\ref{fig:outcomes_above} shows the same when initial velocity is above the safe limit, \ie $v(0) > v^{safe}_{max}$.

{\it Discussion.}
When $v(0) \leq v^{safe}_{max}$ the target $\mathcal{FI}$ results are identical to the best case configuration of $\mathcal{MC}$.
This supports the primary claim of this work, \ie deterministic collision avoidance in the presence of obstacle existence detection faults when operating within predictable safety limits.

When $v(0) > v^{safe}_{max}$ the max range at which the obstacle is detected by the safety layer obstacle detection algorithm ($R^O_{max}$) is not guaranteed to be a sufficient distance for the AV to brake to stop.
This results in more collision outcomes for the target result in $\mathcal{FI}$ as compared to the ideal case of $\mathcal{MC}$ in Figure~\ref{fig:outcomes_above}.

Despite the absence of a deterministic guarantee, Safe stops still happen when $v(0) > v^{safe}_{max}$, because
  \ci although not guaranteed, as shown in Figure~\ref{fig:det_eq}, the discrete nature of \lidar beams leads to regions below the linear bound where obstacles may be detected when beyond $R^O_{max}$;
  \cii false positives cause unnecessary early braking which in this case creates additional slack time for collision avoidance;
  \ciii when $d(0)$ is large, the mission layer's actuation commands reduce AV speed to match its planning and control goals, helping avoid collisions.

It should be noted that the safety layer in \ps does not share the mission layer's goal of completing the AV's mission.
Hence, when the mission layer has safety-critical faults, $\mathcal{MC}$ and $\mathcal{FI}$, the mission of the AV, \ie \textit{Safe Pass}, is abandoned.

$v^{safe}_{max}$ is derived from vehicle, algorithm, and sensor parameters, its value can be increased by changing these parameters, \eg larger \lidar sensor array.

\subsection{Safety - Worst Case}
\label{sec:eval_wc}

\begin{figure}[t]
	\centering
	\subfloat[\label{fig:estop_mission}$\mathcal{MO}$]{%
		\includegraphics[width=0.3\linewidth]{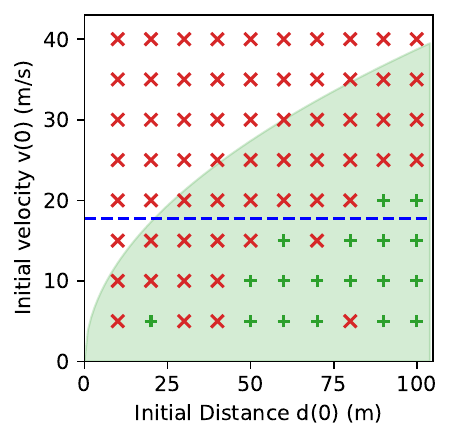}}
	\hfill		
	\subfloat[\label{fig:estop_sr_mission_crash}$\mathcal{MC}$]{%
		\includegraphics[width=0.3\linewidth]{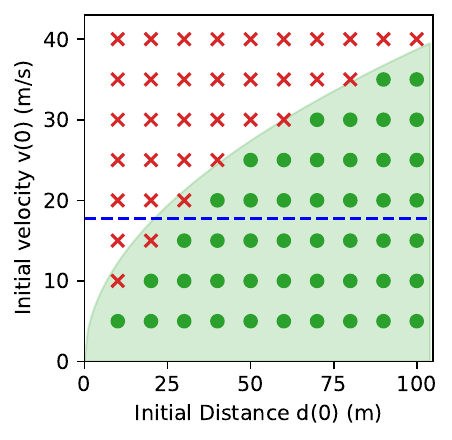}}
	\hfill
	\subfloat[\label{fig:estop_sr_forced_faults}$\mathcal{FI}$]{%
		\includegraphics[width=0.3\linewidth]{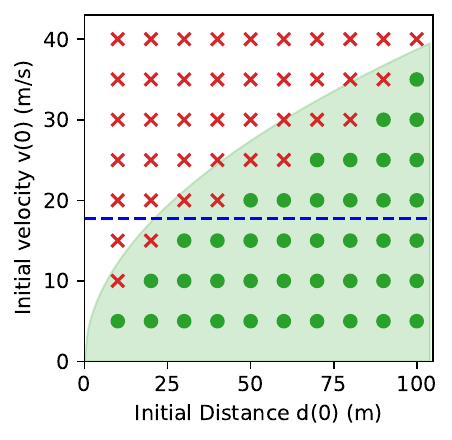}}
	\caption{\label{fig:estop}
		Each data point represents the worst-case outcome among the 10 repetitions conducted for each combination of $d(0)$ (X-Axis), $v(0)$ (Y-Axis), and \ps configuration.
		The shaded green area, defined by $v(0) \leq \sqrt{2*7.5*d(0)}$, represents the distance and velocity combinations at which it is physically possible for the AV to avoid a collision, assuming constant deceleration of $7.5$ $m/s^2$ and zero response latency.
		{\color{red}\textbf{x}} represents collision, \tikzcircle{2pt} represents safe stop and {\color{ForestGreen}\textbf{+}} represents safe pass ($\S$\ref{sec:eval_setup}).
		The blue dashed line notes \mbox{$v^{safe}_{max} = 17.71~m/s$}.
	}
\end{figure}

The simulation results are now presented in higher detail, emphasizing the worst case for each $d(0), v(0)$ and \ps configuration combination,
  as shown in Figure~\ref{fig:estop}.
Among the 10 repetitions for each combination, if any instance results in a collision, the worst-case result notes it as \textit{Collision} ({\color{red}x}).
Only when all repetitions safely pass the obstacle the noted result is a \textit{Safe Pass} ({\color{ForestGreen}\textbf{+}}).
When no collisions occur and at least one instance does not lead to a safe pass, the result is noted to be \textit{Safe Stop} (\tikzcircle{2pt}).

$v^{safe}_{max} = 17.71$ $m/s$ is indicated by blue horizontal lines in each plot.
The shaded area ($v(0) \leq \sqrt{2*7.5*d(0)}$) notes the best possible braking response, assuming $a^{av}_{max} = 7.5$ $m/s^2$.
This ignores any computation latency or dynamics of the simulation.
Figure~\ref{fig:estop_sr_mission_crash} is the synthetic best case for safety where the AV applies maximum brake at the start of the simulation run.

{\it Discussion.}
Figure~\ref{fig:estop_sr_forced_faults} captures the \ps behavior under obstacle existence detection faults, the focus of this work ($\S$\ref{sec:scope}).
This result substantiates the $v^{safe}_{max} = 17.71$ $m/s$ safe operating limit established earlier, as below this limit, the worst case result of the ideal scenario (Figure~\ref{fig:estop_sr_mission_crash}) is same as  Figure~\ref{fig:estop_sr_forced_faults}, the \ps behavior under obstacle existence detection faults.


\subsection{Performance}

Figure~\ref{fig:speeds} shows the comparison of AV speed over time when controlled by the Mission layer only ($\mathcal{MO}$) \textit{vs.} \ps without any instrumented faults.
The evaluation scenario is similar to Figure~\ref{fig:estop_scenario}, except that the obstacle is not in the same lane as the AV, rather it is in an adjacent lane.
Initial AV state is fixed at $d(0) = 50 m, v(0) = 1 m/s$.
$\mathcal{MC}$ and $\mathcal{FI}$ are not evaluated here as they would lead to a safe stop but the mission of the AV will be abandoned due to the instrumented faults.
Compared to the baseline of Mission, \ps shows some additional braking, however erroneous overrides due to transient false positive detections from the safety layer algorithm are mitigated with minimal impact on AV speed, as enforced by Algorithm~\ref{alg:fault_alg}.
The only significant difference near the start ($t \approx 2$) is when the safety layer detects the obstacle whereas the mission layer does not yet.

\begin{table}[t]
	\begin{minipage}[]{0.49\linewidth}
		\centering
		\includegraphics[width=.8\linewidth, trim={0 3mm 0 0}]{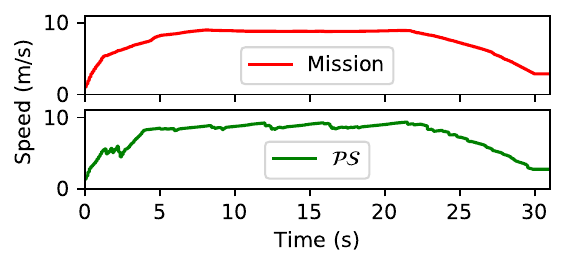}
 		\captionof{figure}{\label{fig:speeds}
			Performance comparison. $d(0)=50m, v(0)=1m/s$.
			At $t=2$ safety layer first detects the obstacle.
		}
	\end{minipage}
	\hfill
	\begin{varwidth}[]{0.49\linewidth}
		\centering
		\caption{\label{tab:comp_lat} Computational Latency}%
		\resizebox{\linewidth}{!}{
			\begin{tabular}{llrr}
				\toprule
				\textit{Layer}	& \textit{Module}     	& \textit{Average}	& \textit{Worst Case}	\\ \midrule
				Mission 		& Segmentation        	& 69362 $\mu$s      & 128634 $\mu$s			\\
				Mission 		& Fusion              	& 922 $\mu$s       	& 3665 $\mu$s			\\ \toprule
				Safety 			& Obstacle Detection	& 7273 $\mu$s       & 38391 $\mu$s			\\ 
				Safety 			& Fault Handlers		& 387 $\mu$s    	& 9801 $\mu$s  			\\ \bottomrule
			\end{tabular}%
		}
	\end{varwidth}%
\end{table}

\subsection{Computational Latency}
\label{sec:eval_latency}

The final result is a comparison of the execution latency of safety and mission layer components.
The results are presented in Table~\ref{tab:comp_lat}.
Safety layer Fault Handler time includes fault detection, criticality determination, and override.
The values only include the computational latency \ie any time spent waiting for inputs is excluded.
Safety layer components have significantly lower latency than those in the mission layer,
  a direct result of the simplified safety requirements and minimalist design of the safety layer components.
The low latency supports a faster response to external stimuli.

\section{Discussion}
\label{sec:discussion}

\textbf{Incremental Design}.
The \ps architecture presented in this work is designed to focus specifically on obstacle existence detection faults due to their prevalence in real-world fatal collisions ($\S$\ref{sec:motivation}). \ps can be expanded to address other perception faults if verifiable algorithms exist that can fulfill the corresponding perception task.
In the absence of such capabilities, safety overrides must be applied precisely to only handled fault types.

Consider the scenario of a deer suddenly jumping in front of an AV using the presented \ps. If the mission layer is unable to detect the existence of the deer, the safety layer overrides the control and applies brakes. This has multiple beneficial effects, such as completely avoiding the collision or reducing the impact velocity if a collision is unavoidable. Additionally, the time before the collision is increased, giving the mission layer more time to react.
When the mission layer does detect the existence of the deer, the safety layer removes its override, even if doing so leads to a collision with the deer. Within the presented design, the safety layer does not have the capability to determine the best possible path for the AV. A collision with the deer can be the best response, avoiding more serious collisions with other vehicles. The safety layer, as presented in this work, does not have the capability to make this distinction and therefore does not intervene.
Therefore, \ps provides guarantees against specific faults only, which can be expanded with the inclusion of other verifiable capabilities. Rather than replacing other prior solutions, the modular design of \ps solves the most critical issue first while remaining open to future expansion and integration.

\noindent \textbf{Trade-offs}.
The improved safe operating capabilities of the AV come at the cost of degradation of rider comfort and performance, \eg due to reduced speed and increased braking (Figure~\ref{fig:speeds}).
\ps minimizes the performance loss by responding to specific faults and only when faults pose a risk of collision.
An advantage of using verifiable and logically analyzable software is that for each hardware and software parameter, the impact of different values can be determined analytically.

\noindent \textbf{Limitations and Future Works}.
There are two main limitations of this work which are also the topic for future works.
First, the presented verification model addresses the main ground vs obstacle differentiation step, however, the complete depth clustering algorithm goes beyond this.
These additional steps reduce the occurrence of false positive detections.
Additionally, as detailed in Section~\ref{sec:constraints} constraints on
  obstacle width (\texttt{\textbf{C2}}) and first point of ground (\texttt{\textbf{C3}}) are minimally restrictive and can be easily mitigated.
However, the inherent limitation of \lidar sensor (\texttt{\textbf{C1}}) cannot be mitigated directly.
Therefore one direction for future work is to create an expanded \textit{detectability} model
unifying different sensor types, all using verifiable algorithms, \ie sensor fusion,
and including the additional algorithmic steps.

Second, while \ps presents safety guarantees in the presence of faults
  without compromising the higher mission layer capabilities when faults are not detected,
  the response to faults is limited to braking and does not leverage the mission layer capabilities.
Therefore the main focus of our future work is to utilize mission layer capabilities in fault tolerance and override actions,
  while maintaining the verifiable safety guarantees presented in this work.

\section{Conclusion}
\label{sec:conclusion}

This work presents \theterm (\ps), a system architecture for fault tolerance against perception faults.
In this work, \ps is applied to the critical issue of obstacle existence detection faults.
The modular design of \ps makes it extensible to cover other types of perception faults, while the verifiable nature of its safety layer ensures that critical safety requirements are met without relying on uncertain machine learning models.
\ps provides a promising direction for the design of fault-tolerant AV. 

\bibliographystyle{unsrturl}
\bibliography{ref}

\end{document}